\documentclass[a4paper, onecolumn]{article}

\usepackage[utf8]{inputenc}\DeclareUnicodeCharacter{2212}{-}
\usepackage{biblatex}
\usepackage[affil-it]{authblk}
\usepackage[a4paper, textwidth=16cm, textheight=25cm]{geometry}
\usepackage{amsmath}
\usepackage{amssymb}
\usepackage{amsfonts} 
\usepackage{pifont} 
\usepackage{nicefrac} 
\usepackage[ruled,vlined]{algorithm2e} 
\usepackage{pgf} 
\usepackage{graphicx,psfrag,color}
\usepackage{float} 
\usepackage{tabularx}
\usepackage{rotating}
\usepackage{booktabs} 
\usepackage{listings}
\usepackage{bigints}
\usepackage{subfigure}
\usepackage{multirow}

\usepackage{tikz} 

\title{\bf DA-VEGAN: Differentiably Augmenting VAE-GAN for microstructure reconstruction from extremely small data sets}
\author[1]{Yichi Zhang}
\author[1]{Paul Seibert}
\author[1]{Alexandra Otto}
\author[1]{Alexander Ra{\ss}loff}
\author[1,2]{Marreddy Ambati}
\author[1,3]{Markus K\"astner\thanks{{Corresponding Author. \it{E-mail address}:} markus.kaestner@tu-dresden.de (Markus K\"astner)}}

\affil[1]{Institute of Solid Mechanics, TU Dresden,  01062 Dresden, Germany}
\affil[2]{Material Mechanics and Durability, GE Global Research, 1 Research Circle, Niskayuna, New York 12309, USA}
\affil[3]{Dresden Center for Computational Materials Science, TU Dresden, 01062 Dresden, Germany}

\date{}
\begin{document}
\maketitle

\hrule

\begin{abstract}
  Microstructure reconstruction is an important and emerging field of research and an essential foundation to improving inverse computational materials engineering (ICME). Much of the recent progress in the field is made based on generative adversarial networks (GANs). Although excellent results have been achieved throughout a variety of materials, challenges remain regarding the interpretability of the model's latent space as well as the applicability to extremely small data sets. The present work addresses these issues by introducing DA-VEGAN, a model with two central innovations. First, a $\beta$-variational autoencoder is incorporated into a hybrid GAN architecture that allows to penalize strong nonlinearities in the latent space by an additional parameter, $\beta$. Secondly, a custom differentiable data augmentation scheme is developed specifically for this architecture. The differentiability allows the model to learn from extremely small data sets without mode collapse or deteriorated sample quality. An extensive validation on a variety of structures demonstrates the potential of the method and future directions of investigation are discussed. \\
  {\bf Keywords}: Microstructure, Reconstruction, GAN, VAE
\end{abstract} 
\hrule

\section{Introduction}
\label{sec:introduction}
In many engineering applications, the underlying materials are random heterogeneous media where the material properties are strongly influenced by the microstructure.
Therefore, materials design offers a great potential in improving the performance of engineering applications and thus facilitates progress across various applications. 
This motivates data-centric approaches such as inverse computational materials engineering (ICME)~\cite{chen_data-centric_2022} or materials-by-design~\cite{choi_artificial_2022} to accelerate progress in materials science.
Therein, microstructure reconstruction, in the following referred to as reconstruction, plays an important role in facilitating \emph{(i)} the generation of several microstructure realizations from few examples, \textit{(ii)} the construction of 3D volume elements from 2D slices and \textit{(iii)} the morphologically meaningful interpolation between microstructures. 
Comprehensive overviews are given in~\cite{bostanabad_computational_2018,bargmann_generation_2018,sahimi_reconstruction_2021}.
The great variety of microstructure characterization and reconstruction (MCR) methods can be separated into the more classical \emph{descriptor-based} approaches and the more recent \emph{data-based} methods that are based on machine learning (ML).

In \emph{descriptor-based} methods, microstructure descriptors are explicitly prescribed.
These descriptors can range from simple volume fractions to Minkowski tensors~\cite{scheunemann_design_2015} or spatial correlations~\cite{yeong_reconstructing_1998,gerke_description_2012,tahmasebi_multiple-point_2012}. 
A well-known method is the Yeong-Torquato algorithm~\cite{yeong_reconstructing_1998}, which is very simple and elegant but computationally demanding for high-resolution 3D structures even after over~20 years of hardware developments due to its stochastic nature.
Common ways to improve its efficiency are to simplify the microstructure morphology~\cite{scheunemann_design_2015} or to decrease the cost per iteration, e.g., by updating descriptors instead of recomputing them from scratch~\cite{adam_efficient_2022}.
Alternatively, reconstruction can be restricted to differentiable descriptors, so that gradient-based optimizers can be used.
This is formulated as \emph{differentiable MCR} in~\cite{seibert_reconstructing_2021-1,seibert_descriptor-based_2022}. Several works can be identified as special cases of this strategy~\cite{li_transfer_2018,bostanabad_reconstruction_2020,bhaduri_efficient_2021}.
An open-source implementation of these approaches is available in \emph{MCRpy}~\cite{seibert_microstructure_2022}.
A less general but extremely efficient alternative is to directly describe the underlying random field. Examples are weakly non-uniform Gaussian random fields~\cite{liu_direct_2017,liu_translation_2016} and explicit mixture random fields~\cite{gao_ultraefficient_2022}.
As a recent contribution to this type of methods, an approach has been proposed to reconstruct anisotropic structures from two-point correlations using multi-output Gaussian random fields~\cite{robertson_efficient_2021}.
Finally, although mainly intended for geometric inclusions and grain-like structures, \emph{DREAM.3D}~\cite{groeber_dream3d_2014} is arguably by far the most often applied descriptor-based MCR tool in practice.

In \emph{data-based} methods, the explicit microstructure descriptor is replaced by a latent representation that is learned from data.
Although this latent representation can be interpreted as a descriptor, these approaches are special in that the representation itself is learned from a data set.
An early example is non-parametric resampling, introduced to the MCR community in~\cite{bostanabad_stochastic_2016} and still developed further today~\cite{latka_microstructure_2021}.
New trends include transformers~\cite{phan_size-invariant_2022} and diffusion models~\cite{dureth_conditional_2023,lee_microstructure_2023}.
However, most of the recent research has been focused on reconstruction using generative adversarial networks (GANs) and variational autoencoders (VAEs).
These methods are discussed in Section~\ref{sec:literature}.
In general, data-based methods outperform descriptor-based methods in that complex microstructures can be reconstructed much faster\footnote{Random field based methods are also extremely efficient but all current approaches are restricted to structures with limited morphological complexity.}: While the training phase is generally computationally expensive, the reconstruction itself only requires to query the trained model and can be performed in a matter of seconds or milliseconds.
However, latent spaces of neural networks often exhibit strong gradients~\cite{verma_manifold_2019}, so that a very small change in the latent space can lead to a completely different microstructure.
This affects applications such as the establishment of structure-property linkages and hinders controlled synthesis, i.e., the  targeted variation of certain microstructural features and the construction previously non-existent microstructures with desired characteristics~\cite{nguyen_synthesizing_2022}.
This is a disadvantage of latent spaces over classical descriptors like spatial correlations, which are generally smoother, making them easier to interpret and more suitable for learning structure-property linkages.
The biggest disadvantage of data-based methods, however, is the need for a (potentially large) micrograph data set, which is extremely time- and cost-intensive to generate.
These two problems are also identified as major challenges for materials-by-design in a recent review~\cite{choi_artificial_2022}.

This work aims at addressing these disadvantages of GAN-based reconstruction algorithms.
A GAN-based architecture is chosen, where the latent space is provided by a $\beta$-VAE.
As opposed to conventional VAEs, a $\beta$-VAE introduces a parameter~$\beta$ to control the trade-off between a good reconstruction and a smooth latent space.
Furthermore, a differentiable data augmentation scheme is incorporated in order to handle even extremely small data sets.
The proposed model is therefore called DA-VEGAN as a backronym for differentiably augmenting variational autoencoder and generative adversarial network.

After a brief review of GANs and GAN-based reconstruction in Section~\ref{sec:literature}, DA-VEGAN is presented in Section~\ref{sec:methods}.
A thorough validation is performed in Section~\ref{sec:numericalexperiments} and a conclusion is drawn in Section~\ref{sec:summary}.

\section{Brief introduction to GAN- and VAE-based microstructure reconstruction}
\label{sec:literature}
A generative adversarial network (GAN) is a machine learning framework where two adversarial models improve each other by standing in direct competition~\cite{goodfellow_generative_2014}.
Specifically, the \emph{first} model, the generator~$G(z)$ maps a random noise vector~$z$ to a synthetic data point or image~$x$, whereas the \emph{second} model, the discriminator~$D(x)$ predicts whether~$x$ stems from the data set or is synthetic.
Given a training data set, a GAN learns to generate data points that are not in the training data set, but closely resemble points therein.
Even on complex image data such as human portraits, remarkable results can be achieved~\cite{karras_analyzing_2019}.

In the context of MCR, to the authors' best knowledge, the first application of GANs dates back to 2017~\cite{mosser_reconstruction_2017}.
After that, a great variety of architectures and extensions was developed and applied to numerous materials.
Despite the abundance of methods, no review paper specifically for GAN-based MCR is known to the authors.
The latest review article on microstructure reconstruction~\cite{sahimi_reconstruction_2021} covers the developments until 2021 and a brief summary is given in the following.
In face of the adversarial training of GANs, enhanced training stability is an important area of research, where additional invariances (such as rotation and symmetry) are considered as well as robust and progressively growing architectures~\cite{xia_multi-scale_2022,henkes_three-dimensional_2022}.
Furthermore, the GAN modifications developed by the ML community are assessed regarding their merit for microstructural data.
Examples include conditional GANs~\cite{shams_hybrid_2021} for steering the reconstruction by labels and BiCycle GANs~\cite{feng_end--end_2020} for preventing mode collapse.
Besides that, GANs are combined with other ML models to obtain hybrid models.
For example, recurrent neural networks that operate in the latent space of a 2D encoder-decoder architecture provide an interesting extension to reconstruct 3D structures from 2D data~\cite{zhang_3d-pmrnn_2022}. However, it should be mentioned that GANs alone can also achieve 2D-to-3D reconstruction by means of an adapted training procedure~\cite{kench_generating_2021,li_cascaded_2022}.
Finally, the success of transformer models in ML~\cite{vaswani_attention_2017} motivated their application to microstructure reconstruction~\cite{zheng_rockgpt_2022}.
In this context, combinations of GANs and transformers have been shown to achieve excellent results~\cite{phan_size-invariant_2022}.

Autoencoders, however, are the most frequently chosen model for a hybrid GAN architecture. 
An autoencoder is a machine learning model that compresses high-dimensional input data~$x$ to a lower-dimensional latent vector~$z$ and then regenerates~$x$ from~$z$~\cite{doersch_tutorial_2021}.
Thus,~$z$ is learned such that it defines a non-linear manifold that accurately captures the training data set.
As a special case of these models, variational autoencoders (VAEs) enforce the relation between input image and latent space to be smooth and to have small gradients by, in simple words, introducing random perturbations to~$z$~\cite{doersch_tutorial_2021}.
This facilitates a smooth transition between two points in the latent space.
Furthermore, if a small change in the latent space only corresponds to a small change in the input or generated image, then the latent space might be more interpretable than if strong nonlinearities are present.
Motivated by this observation, $\beta$-VAEs introduce a hyperparameter~$\beta$ to trade off the image reconstruction quality against the robustness of the latent space~\cite{doersch_tutorial_2021}.
In summary, VAEs allow to define suitable latent spaces, however, the quality of the generated images cannot compete with that of a GAN model~\cite{bond-taylor_deep_2022}.

Autoencoders alone are applied in~\cite{cang_microstructure_2017} and modifications like batch normalization are investigated later in~\cite{zhang_3d_2022}. 
An application to real materials such as batteries~\cite{faraji_niri_performance_2022} shows promising results, however, a combination of GAN and autoencoder promises further improvements.
The ability of autoencoders, or specifically VAEs, to define a suitable latent space for a given data distribution in combination with the high quality achieved by GANs has led to excellent results~\cite{shams_coupled_2020,zhang_slice--voxel_2021}
Moreover, as opposed to a GAN-only architecture, the trained hybrid model comprises an encoder that maps from data to the latent space.
Given a microstructure~$x$, this allows to use the latent space~$z$ as a computable microstructure descriptor and to assess the quality of the model by comparing the generator output~$G(z)$ with~$x$, which is not possible for a pure GAN architecture.

\section{Proposed approach}
\label{sec:methods}
The hybrid architecture of DA-VEGAN and the corresponding loss functions are presented in Sections 3.1 and 3.2, respectively.
Finally, a differentiable data augmentation scheme is presented in Section 3.3.

\subsection{Hybrid architecture}
\label{sec:methods_hybridarchitecture}
Combining VAEs and GANs into one single architecture aims at harnessing both their advantages.
For this reason, a single network serves as both generator for the GAN and decoder for the VAE to integrate the two models into one hybrid architecture as shown in Figure~1. 
This opens up the possibility to create a smoother and more interpretable latent space with help of the encoder while improving the general quality of the synthetic microstructures by an adversarial training methodology.

Regarding a given set of microstructures as a multi-variate prior distribution, the encoder defines an inferred posterior distribution in the form of a mean vector~$\mu$ and its corresponding variance vector~$\sigma$. 
In our proposed model, the encoder learns the mean and variance vectors together as a combined vector that is later split into mean and variance to define a latent distribution. 
Following common practice, a latent vector $z_{\text{vae}}$ is sampled from this distribution, while gradients are defined by reparametrization.
Furthermore, latent vectors $z_{\text{noise}}$ are sampled directly from a multivariate standard normal distribution to increase the diversity and quality of the generated microstructures.

Given these latent vectors, the two goals of the mutual generator are to \emph{(i)} reconstruct the pixel-based microstructure representation~$x$ from the latent vector that matches the original sample~$x_{\text{real}}$ from the training data set as precise as possible, and \emph{(ii)} simultaneously generate synthetic microstructures from random latent vectors that are indistinguishable from the training data to the discriminator. 
The outputs of the generator $G(z_{\text{vae}})$, $G(z_{\text{noise}})$ are augmented separately by the differentiable data augmentation function $T(x)$, which is introduced in following sections.

Unlike with a purely GAN-based architecture, the microstructures generated from $z_{\text{vae}}$ should also be labeled as "real" if they are precise reconstructions from the training data set. 
Thus, the input of the discriminator includes not only the training data set and $G(z_{\text{noise}})$, but also $G(z_{\text{vae}})$. 
The predictions of the discriminator are then evaluated by various loss functions outlined in Section 3.2.
An optimization is applied to train the generator and discriminator respectively. 
In contrast, the encoder is concurrently trained with a custom loss function.
This is discussed in greater detail in Section 3.3.2 together with the corresponding differentiable data augmentation scheme.

\subsection{Loss functions}
\label{sec:loss_functions}
The three-component architecture comprising an encoder, a discriminator and a generator, motivates the definition of three distinct loss functions $\mathcal{L}_{\text{Enc}}$, $\mathcal{L}_{\text{Disc}}$ and~$\mathcal{L}_{\text{Gen}}$, respectively.
During training, the gradients with respect to the parameters of a model are only taken from its corresponding loss function.
Consequently, model coupling is not realized by means of cross-derivatives, but by choosing the loss functions accordingly.

Following common practice for VAEs, the loss function of the encoder
\begin{equation}
    \mathcal{L}_{\text{Enc}} = \beta \cdot \mathcal{L}_{\text{KLD}} + \mathcal{L}_{\text{Rec}} \;
\end{equation}
is given as weighted sum of the Kullback-Leibler~(KL) divergence term, also called regularization term
 \begin{equation}
\mathcal{L}_{\text{KLD}} = D_{\text{KL}} \bigl( q_{\Theta}(z|x) || p(z) \bigr) \;
\end{equation}
and the reconstruction term
 \begin{equation}
\mathcal{L}_{\text{Rec}} = -\log \bigl( p_{\Phi}(x|z) \bigr) \; ,
\end{equation}
which evaluate the regularization error and the pixel wise reconstruction error,  respectively.
Herein,~$x$ and~$z$ denote the natural pixel-based and latent representation of a microstructure, respectively.
The scalar parameter~$\beta$ for the regularization allows to penalize strong gradients in the latent space.
Furthermore,~$q_{\Theta}(z|x)$ is the approximation of the posterior distribution achieved by the encoder parameterized with~$\Theta$, and~$p_{\Phi}(x|z)$ is the reconstruction of the original input by the mutual generator/decoder parameterized with~$\Phi$. 
Finally, the definition of the Kullback-Leibler divergence~$D_{\text{KL}}$ deserves further attention:
In contrast to the common form of~$D_{\text{KL}}$, the differentiable VAE data augmentation developed in this work requires a more general formulation, as discussed in Section 3.3.2.

The loss functions of the generator and discriminator are chosen by building upon the typical GAN formulation for the generator 
\begin{equation}
    \Tilde{\mathcal{L}}_{\text{Gen}} = \log \Bigl(1 - D \bigl( G(z) \bigr) \Bigr)
\end{equation}
and the discriminator
\begin{equation}
    \Tilde{\mathcal{L}}_{\text{Disc}} = -\log \bigl( D(x) \bigr) - \log \Bigl(1 - D \bigl( G(z) \bigr) \Bigr) \; .
\end{equation}

For the discriminator, beyond Equation (5), decoded samples should not only be evaluated from random latent states $z_{\text{noise}}$ (synthetic samples), but also from the latent representations given by the encoder $z_{\text{vae}}$ (reconstructed versions of real samples).
Hence, the corresponding loss function is defined as
\begin{equation}
    \mathcal{L}_{\text{Disc}} = -\log \bigl( D(x) \bigr) - \log \Bigl(1 - D \bigl( G(z_{\text{vae}}) \bigr) \Bigr) - \log \Bigl( 1 - D \bigl( G(z_{\text{noise}}) \bigr) \Bigr) \; .
\end{equation}

The same applies to the loss function of the generator
\begin{equation}
     \mathcal{L}_{\text{Gen}} = -\lambda_{\text{vae}} \cdot \log \Bigl( D \bigl( G(z_{\text{vae}}) \bigl) \Bigr) - \lambda_{\text{noise}} \cdot \log \Bigl( D \bigl( G(z_{\text{noise}}) \bigr) \Bigr) + \lambda_{\text{rec}} \cdot \mathcal{L}_{\text{Rec}} \; ,
\end{equation}
which contains two discriminator-based loss terms for microstructures generated from $z_{\text{vae}}$ and $z_{\text{noise}}$, respectively.
In this context, it is pointed out that the typical term $\log (1-D)$ from Equation (4) is replaced by $-\log D$, which is known to enhance training stability in some cases~\cite{gulrajani_improved_2017}.
Finally, the reconstruction term~$\mathcal{L}_\text{Rec}$ in Equation~(3) is added to $\mathcal{L}_{\text{Gen}}$.
This makes the generator fulfil its role as decoder in the hybrid architecture.
It is worth noting that the generator, much like a decoder in a conventional VAE, is merely responsible  for reconstructing $x$ from $z$, whereas the definition of $z$ is exclusively given by the encoder~\cite{yu2020tutorial}. 
Therefore, the mutual generator in the hybrid architecture does not require to minimize the regularization loss (2).

This approach constitutes the state of the art in 2D-to-2D reconstruction, whereas more complex loss functions can be used to demand specific features of the generated microstructures or to reconstruct 3D structures from 2D training data.
The approach by Zhang et al.~\cite{zhang_slice--voxel_2021} is based on a similar architecture and it might be promising to be transferred to our model.
However, the extension to 3D is left open for future investigations, since this work is focused on smooth latent spaces and small data sets.

\subsection{Differentiable data augmentation}
\label{sec:methods_dataaugmentation}
\subsubsection{GAN augmentation}

As already addressed in the introduction, the performance of GAN-based reconstruction relies heavily on a sufficient amount of diverse and high-quality training samples, which is often prohibitively expensive. 
A typical strategy to reduce the over-fitting caused by insufficient training data is data augmentation.
It aims at significantly increasing the diversity of the training data set from a limited amount of existing samples.
However, traditional augmentations including cropping, flipping, scaling, brightness altering and cutout make the generator match the augmented distribution instead of the original~\cite{zhao2020differentiable}.
The same applies to data augmentation by means of an auxiliary black-box microstructure reconstruction approach, which is sometimes used in the literature~\cite{kamrava_end--end_2022,lee_microstructure_2022}.

Neural networks are mostly optimized via gradient-based methods.
In order to eliminate the influence of the augmentation, gradients caused by a data augmentation should also be able to be back-propagated.
This motivates the idea of a differentiable data augmentation on GAN architectures to resolve such difficulties.
In the work of Zhao et al.~\cite{zhao2020differentiable}, augmentations are implemented as differentiable functions in the training process. 
By augmenting, both, the training data set as well as the generated pictures, the model generates excellent results even with reduced training data. Various combinations of differentiable augmentations have been tested and have shown to outperform traditional methods for GAN architectures, especially with small data sets. 

Since the training data in this work is assumed to consist of segmented multi-phase microstructures, many common augmentation strategies such as slight re-coloring or cropping and zooming cannot be applied as they would result in significantly altered or invalid structures.
Similarly, image rotation is deliberately excluded in view of the relevance of the orientation of microstructural features.
This leaves the augmentations restricted to translation only. 
The translation augmentation can be described as a displacement of the entire microstructure, where the entries of the displacement vector are sampled from a normal distribution. 
To avoid a massive loss of information after long-range translation, periodic boundaries are used.
This introduces a small error, because the original micrographs in the training data set are naturally not periodic, however, the differentiability of this augmentation ensures that this does not affect the quality of the generated samples.
\begin{figure}[t] 
    \centering
    \begin{tikzpicture}[scale=0.5]
        \node [draw, align=left, rounded corners] at (1.8,-4.5) {\small Gauss Noise};
        \node [draw, rounded corners] at (2, -7) {\large $x_{\text{real}}$};

        \filldraw[fill=green!25]
                (0,-2)
                --(0,2)
                --(4,1)
                --(4,-1)
                --cycle;
        \node [] at (2,0.4) {\small Encoder};
        \node [] at (2,-0.5) {$E(x)$};
        
        \begin{scope}[shift={(7.5,-0.5)}]
            \filldraw[fill=red!25]
                (0,-1)
                --(0,1)
                --(4,2)
                --(4,-2)
                --cycle;
            \node [] at (2,0.4) {\small Generator};
            \node [] at (2,-0.5) {$G(z)$};
            \draw[-latex, thick]
                (4,0.7)
                --(4.9,0.7);
            \draw[-latex, thick]
                (4,-0.7)
                --(4.9,-0.7);
        \end{scope}
        
        \node [draw, rounded corners] at (13.6,0.2) {\small $G(z_{\text{vae}})$};
        \node [draw, rounded corners] at (13.8,-1.2) {\small $G(z_{\text{noise}})$};

        \filldraw[fill=gray!25, rounded corners]
            (4.8,-1.6)
            --(4.8,1.6)
            --(6.3,1.6)
            --(6.3,-1.6)
            --cycle;
        \node [] at (5.55,0) {\small $z_{\text{vae}}$};
        \begin{scope}[shift={(0,-4.5)}]
            \filldraw[fill=gray!25, rounded corners]
                (4.8,-1.6)
                --(4.8,1.6)
                --(6.3,1.6)
                --(6.3,-1.6)
                --cycle;
        \node [] at (5.55,0) {\small $z_{\text{noise}}$};
        \end{scope}
        
        \begin{scope}[shift={(18,-6)}]
            \filldraw[fill=cyan!50!gray]
                (0,-2)
                --(0,2)
                --(4,1)
                --(4,-1)
                --cycle;
            \node [] at (2,0.4) {\small Discriminator};
            \node [] at (2,-0.5) {$D(x)$};

            \draw[-latex, thick]
                (4,0)
                --(8.5,0)
                --(8.5,6.8)
                --(7,6.8);
            \draw[-latex, thick]
                (8.5,3.8)
                --(8.5,9)
                --(7.7,9);
            
            \node [draw, align=left, rounded corners] at (5,9) { $D(T(G(z_{\text{vae}})))$\\  $D(T(G(z_{\text{noise}})))$};
            \node [draw, align=left, rounded corners] at (5,6.8) { $D(T(x_{\text{real}}))$};

            \draw[-latex, dashed, thick]
                (2.3, 9.4)
                --(-8.5, 9.4)
                --(-8.5, 7);
            \node[] at (-3.5, 10) {$\mathcal{L}_\mathrm{Gen}$};
            
            \draw[-latex, dashed, thick]
                (2.3, 8.6)
                --(1.5,8.6)
                --(1.5,1.6);
            \draw[dashed, thick]
                (2.9, 6.8)
                --(1.5, 6.8);
            \node[] at (4, 3.8) {$\mathcal{L}_\mathrm{Disc}$};

        \end{scope}
        
        \draw[-latex, thick]
            (4,0)
            --(4.8,0);
        \draw[-latex, thick]
            (6.3,0)
            --(7.5,0);
        
        \draw[-latex, thick]
            (3.8,-4.5)
            --(4.8,-4.5);
        \draw[-latex, thick]
            (6.3,-4.5)
            --(6.8,-4.5)
            --(6.8,-1)
            --(7.5,-1);
        
        \draw[-latex, thick]
            (2.95,-7)
            --(8.2,-7);
        \begin{scope}[shift={(9., -7)}]
            \filldraw[fill=yellow!80!gray, rounded corners]
                (-1.5,-.6)
                --(-1.5,.6)
                --(1.5,.6)
                --(1.5,-.6)
                --cycle;
            \node [] at (0.,0.) {$T(x_{\text{real}})$};
        \end{scope}
        \draw[-latex, thick]
            (10.5,-7)
            --(18,-7);

        \draw[-latex, thick]
            (1.05,-7)
            --(-1,-7)
            --(-1,0)
            --(0, 0);
        
        \draw[-latex, thick]
            (14.9,0.2)
            --(16,0.2)
            --(16,-3);
        
        \draw[-latex, thick]
            (16,-3.5)
            --(16,-5)
            --(18,-5);
        
        \begin{scope}[shift={(16, -3)}]
            \filldraw[fill=yellow!80!gray, rounded corners]
                (-1.85,-.6)
                --(-1.85,.6)
                --(1.85,.6)
                --(1.85,-.6)
                --cycle;
            \node [] at (0.,0.) {$T(G(z_{\text{vae}}))$};
        \end{scope}
            
        \draw[-latex, thick]
            (13.5,-1.7)
            --(13.5,-6);
        \begin{scope}[shift={(13.5, -6)}]
            \filldraw[fill=yellow!80!gray, rounded corners]
                (-2.2,-.6)
                --(-2.2,.6)
                --(2.2,.6)
                --(2.2,-.6)
                --cycle;
            \node [] at (0.,0.) {$T(G(z_{\text{noise}}))$};
        \end{scope}
        \draw[-latex, thick]
            (15.7,-6)
            --(18,-6);

    \end{tikzpicture}
    \caption[Display of the modified hybrid architecture]{Display of the modified hybrid architecture: Real samples~$x_\mathrm{real}$ are encoded as latent variables~$z_\mathrm{vae}$ by the encoder~$E$. The generator~$G$ creates microstructures from both,~$z_\mathrm{vae}$ and randomly sampled~$z_\mathrm{noise}$ that are contrasted to training data by the discriminator~$D$. The output is used to train~$G$ and~$D$, whereas the VAE training is discussed in Figure~2. All inputs to~$D$ are subjected to a differentiable augmentation~$T$ to make the model applicable to extremely small data sets.} 
\end{figure}
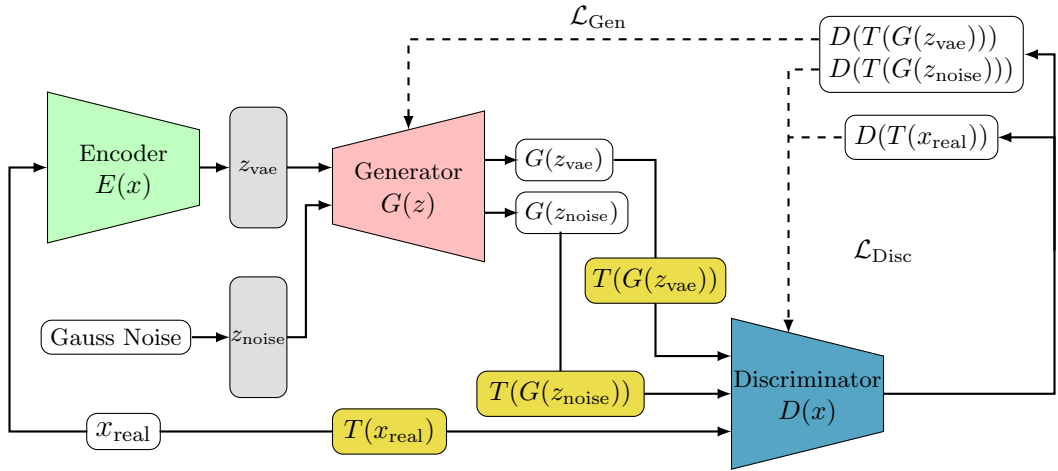

\subsubsection{VAE augmentation}
Even though the data augmentation scheme given in Section 3.3.1 alone significantly enhances the results of the generative model, it cannot improve the performance of the VAE architecture. 
More specifically, the encoder is still heavily dependent on abundant training data to generate meaningful latent variables approaching the target distribution. 
Similarly to how a GAN can be augmented by translations directly in the pixel space, a VAE can be augmented in the latent space.
Because the VAE is characterized by a smooth latent space, a small perturbation of~$z$ represents a similar structure and can be used as an augmentation, as outlined in the following.

A VAE loss function is typically divided into a regularization term and a reconstruction term as given in Equation (1). 
The reconstruction term can be regarded as a direct comparison between the reconstructed and original microstructure.
Therefore, this term is not affected by the GAN augmentation scheme. 
However, the regularization term deserves further attention.

The regularization error is defined as the Kullback-Leibler~(KL) divergence. 
With $\mu$ as mean and $\sigma$ as variance, the definition of the KL-loss is given as
\begin{equation}
D_{\text{KL}} = \mathrm{log}\left( \frac{\sigma_\mathrm{prior}}{\sigma_\mathrm{post}} \right) + \frac{\sigma_\mathrm{post}^2 + (\mu_\mathrm{prior} - \mu_\mathrm{post})^2}{2 \cdot \sigma_\mathrm{prior}^2} - \frac{1}{2} \; ,
\end{equation}
where the subscripts prior and post refer to the prior distribution and the posterior distribution generated from the encoder, respectively. 
The typical form of the KL-loss 
\begin{equation}
\tilde{D}_{\text{KL}} = -\frac{1}{2} \cdot \bigl(1 + \mathrm{log}(\sigma_\mathrm{post}^2) - \sigma_\mathrm{post}^2 - \mu_\mathrm{post}^2 \bigr) \; , 
\end{equation}
which is commonly implemented in VAE loss functions is a special case in which the prior follows a Gaussian normal distribution with $\mu_\mathrm{prior} = 0$ and $\sigma_\mathrm{prior} = 1$.

To increase the diversity of the training data set, the differentiable VAE augmentation functions $T_{\mu}(\mu, u_\mu)$ and $T_{\sigma}(\sigma, u_\sigma)$ are implemented in the loss function to modify the mean and variance respectively, where $u_\mu$ and~$u_\sigma$ are randomized parameters sampled in the augmentation from uniform distributions between~0 and~1.
Augmenting $\sigma_\mathrm{post}$ and $\mu_\mathrm{post}$ directly has shown to improve the performance of the encoder only slightly.
In this context, it is noted that the importance of modifying both the "real" and "fake" sets is repeatedly emphasized the work of Zhao et al. ~\cite{zhao2020differentiable}. 
Therefore, instead of using the specialized form in Equation~(9), the general form of the KL divergence in Equation~(8) is used as the regularization term in the VAE loss function, where $\sigma_\mathrm{prior}$ and $\mu_\mathrm{prior}$ are sampled from a Gaussian normal distribution and can be regarded as "real" input, while $\sigma_\mathrm{post}$ and $\mu_\mathrm{post}$ can be regarded as "fake" input.

The generator augmentation function $T(x, u)$ translates microstructures according to a 2D displacement vector $u=[u_{x}, u_{y}]$. 
Since generators in GANs are encouraged to generate diverse data as long as they are classified as "real" by the discriminator, there is no incentive to prevent microstructural features from being displaced to a different location.
However, the VAE loss function directly evaluates the pixel-wise difference between the input and output, opposed to the GAN loss functions that evaluate the classification results of the discriminator. 
In order to minimize the influences of VAE augmentations, its parameters are shared between augmentations for the same type of variables, namely means, variances and microstructures.
\begin{figure}[ht]
    \centering
    \begin{tikzpicture}[scale=0.5]
        \node [draw, align=left, rounded corners] at (-2.2,6.5) {\small Gauss Noise};
        \node [draw, rounded corners] at (-4.2, -5) {\large $x_{\text{real}}$};
        \filldraw[fill=green!25]
            (0,-2)
            --(0,2)
            --(4,1)
            --(4,-1)
            --cycle;
        \node [] at (2,0.4) {\small Encoder};
        \node [] at (2,-0.5) {$E(x)$};
        \draw[-latex, thick]
            (4,0)
            --(4.8,0);
        
        \filldraw[fill=gray!25, rounded corners]
            (4.8,-1.6)
            --(4.8,1.6)
            --(6.3,1.6)
            --(6.3,-1.6)
            --cycle;
        \node [] at (5.55,0.6) {\small $\sigma_{\mathrm{post}}$};
        \node [] at (5.55,-0.6) {\small $\mu_{\mathrm{post}}$};
        \draw[-latex, thick]
            (6.3,0)
            --(7.8, 0)
            --(7.8, 1.5)
            --(9.2,1.5);
        \draw[-latex, thick]
            (7.8,0)
            --(7.8,-1.5)
            --(9.2,-1.5);
        \begin{scope}[shift={(5.1, 2.2)}]
            \filldraw[fill=yellow!80!gray, rounded corners]
                (4.0,-1.5)
                --(4.0,1.5)
                --(7.3,1.5)
                --(7.3,-1.5)
                --cycle;
        \node [] at (5.65,0.6) {\small $T_{\sigma}(\sigma_{x}, u_{\sigma})$};
        \node [] at (5.65,-0.6) {\small $T_{\mu}(\mu_{x}, u_{\mu})$};
        \end{scope}
        \begin{scope}[shift={(5.1,-1.5)}]
            \draw[rounded corners]
                (4.1,-0.5)
                --(4.1,0.5)
                --(7,0.5)
                --(7,-0.5)
                --cycle;
        \node [] at (5.55,0) {$z_{\text{vae}}$};
        \end{scope}
        \draw[-latex, thick]
            (12.1,-1.5)
            --(14.5, -1.5);

        \begin{scope}[shift={(14.5,-1.5)}]
            \filldraw[fill=red!25]
                (0,-1)
                --(0,1)
                --(4,2)
                --(4,-2)
                --cycle;
            \node [] at (2,0.4) {\small Generator};
            \node [] at (2,-0.5) {$G(z)$};
            \draw[-latex, thick]
                (4,0)
                --(5,0)
                --(5,-5.5)
                --(4,-5.5);
            \draw[rounded corners]
                (4,-4.9)
                --(4,-6.1)
                --(1, -6.1)
                --(1,-4.9)
                --cycle;
            \node [] at (2.5, -5.5) {$G(z_{\text{vae}})$};
            \draw[-latex, thick]
                (1,-5.5)
                --(-1.5,-5.5);
        \end{scope}

        \draw[-latex, thick]
            (-0.2,6.5)
            --(0.8,6.5);
        \begin{scope}[shift={(-4,6.5)}]
            \filldraw[fill=gray!25, rounded corners]
                (4.8,-1.6)
                --(4.8,1.6)
                --(6.4,1.6)
                --(6.4,-1.6)
                --cycle;
            \node [] at (5.6,0.6) {\small $\sigma_{\mathrm{prior}}$};
            \node [] at (5.6,-0.6) {\small $\mu_{\mathrm{prior}}$};
            
            \draw[-latex, thick]
                (6.4,0)
                --(8.1,0);
        \end{scope}
        
        \begin{scope}[shift={(0.1,6.5)}]
            \filldraw[fill=yellow!80!gray, rounded corners]
                (3.5,-1.6)
                --(3.5,1.6)
                --(7.6,1.6)
                --(7.6,-1.6)
                --cycle;
            \node [] at (5.55,0.6) {\small $T_{\sigma}(\sigma_{\mathrm{prior}}, u_{\sigma})$};
            \node [] at (5.55,-0.6) {\small $T_\mu(\mu_{\mathrm{prior}}, u_{\mu})$}; 
        \end{scope}
        
        \begin{scope}[shift={(0, 3.)}]
            \filldraw[fill=yellow!80!gray, rounded corners]
                (4.1,-0.5)
                --(4.1,0.5)
                --(7,0.5)
                --(7,-0.5)
                --cycle;
            \node [] at (5.55,0) {$u_{\sigma},\text{ } u_{\mu}$};
            \draw[-latex, thick]
                (5.6, 0.5)
                --(5.6, 1.9);
            \draw[-latex, thick]
                (7,0)
                --(9.15, 0);
        \end{scope}

        %
        \begin{scope}[shift={(4,-10.5)}]
            \filldraw[fill=yellow!80!gray, rounded corners]
                (2.,-0.6)
                --(2.,0.6)
                --(9,0.6)
                --(9,-0.6)
                --cycle;
        \node [] at (5.5,0) {$T(x_{\text{real}}, [u_{\text{x}}, u_{\text{y}}])$};
        \end{scope}
        \begin{scope}[shift={(9,-8.75)}]
            \filldraw[fill=yellow!80!gray, rounded corners]
                (4.1,-0.6)
                --(4.1,0.6)
                --(7,0.6)
                --(7,-0.6)
                --cycle;
        \node [] at (5.55,0) {$[u_{\text{x}}, u_{\text{y}}]$};
        \draw[-latex, thick]
            (4.1,0)
            --(2,0)
            --(2,1.2);
        \draw[-latex, thick]
            (4.1,0)
            --(2,0)
            --(2,-1.2);
        \end{scope}

        \begin{scope}[shift={(4,-7)}]
            \filldraw[fill=yellow!80!gray, rounded corners]
                (2.,-0.6)
                --(2.,0.6)
                --(9,0.6)
                --(9,-0.6)
                --cycle;
            \node [] at (5.5,0) {$T(G(z_{\text{vae}}), [u_{\text{x}}, u_{\text{y}}])$};
            \draw[-latex, thick]
                (2.,0)
                --(0,0);
        \end{scope}

        \begin{scope}[shift={(13,5)}]
            \draw[thick]
                (0,0)
                --(4,0)
                --(4,2)
                --(0,2)
                --cycle;
            \node [] at (2,1) {$\mathcal{L}_\mathrm{KLD}$};

        \end{scope}

        \begin{scope}[shift={(0,-8.5)}]
            \draw[thick]
                (0,0)
                --(4,0)
                --(4,2)
                --(0,2)
                --cycle;
            \node [] at (2,1) {$\mathcal{L}_\mathrm{Rec}$};
        \end{scope}
        
        \draw[-latex, thick]
            (-3.2,-5)
            --(-1.5,-5)
            --(-1.5,0)
            --(0,0);

        \draw[-latex, thick]
            (-1.5,-5)
            --(-1.5,-10.5)
            --(6,-10.5);
            
        \draw[-latex, dashed, thick]
            (2,-6.5)
            --(2,-1.5);
        \draw[-latex, dashed, thick]
            (2,-4.5)
            --(16.5,-4.5)
            --(16.5, -3);
            \node [] at (9,-3.8) {$\lambda_{\text{rec}} \cdot \mathcal{L}_{\text{Rec}}$};
        \draw[-latex, dashed, thick]
            (17,6)
            --(19,6)
            --(19,9)
            --(-5,9)
            --(-5,3.5)
            --(2,3.5)
            --(2,1.5);
        \draw[-latex, thick]
            (7.65,6.5)
            --(13,6.5);
        \draw[-latex, thick]
            (10.55,3.725)
            --(10.55,5.5)
            --(13,5.5);
        \draw[-latex, thick]
            (7.5, -9.9)
            --(7.5, -8)
            --(4,-8);
        
    \end{tikzpicture}
    \caption[Display of the modified VAE]{Display of the modified VAE with two distinct augmentations shown in yellow: Real samples~$x_\mathrm{real}$ are encoded as latent variables~$z_\mathrm{vae}$ by the encoder~$E$ as a mean and variance vector, $\mu$ and~$\sigma$, respectively. It is compared to a Gaussian by the Kullback-Leibler~(KL) divergence to regularize~$E$. Furthermore, it is reconstructed by the decoder or generator~$G$ to compute a reconstruction loss to train~$E$ and~$G$. Before the computation of the losses, both, latent vectors as well as microstructures, are subjected to differentiable augmentations~$T_\sigma$, $T_\mu$ or~$T$ to make the model applicable to extremely small data sets. These augmentations are parametrized by~$u_\sigma$, $u_\mu$ and~$u$, respectively, which are drawn at random.} 
\end{figure}
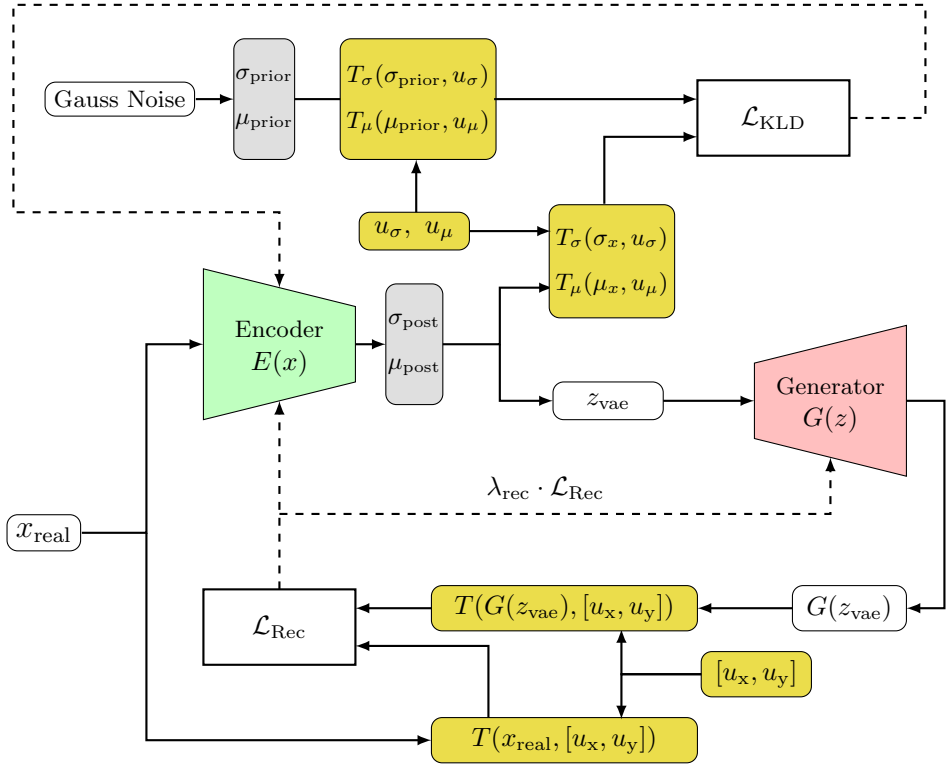

The architecture and the training of the augmented VAE are illustrated in Figure 2. 
Herein, $\sigma_\mathrm{prior}$ and $\mu_\mathrm{prior}$ are mean and variance, which are sampled from a normal distribution and augmented by the augmentation function with parameter sets $u_\mu$ and $u_\sigma$, respectively. 
The same augmentations are applied to $\sigma_\mathrm{post}$ and $\mu_\mathrm{post}$, while the unmodified mean and variance are reparametrized to sample the latent vector $z_{\text{vae}}$. 
The modified "real" as well as "fake" means and variances are then evaluated by the regularization term of the loss function $\mathcal{L}_{\text{KLD}}$ in Equation~(2). 
The same logic applies to evaluating the reconstruction error: The reconstructed microstructures are augmented using the same translation vector $u$ as for the training data set. 
The encoder is optimized directly based on the results of both of the loss functions. 
In contrast, the result of the reconstruction loss function is multiplied by a hyperparameter $\lambda_\mathrm{rec}$ and added to the generator loss function, as explained in Equation (7).

\section{Numerical experiments}
\label{sec:numericalexperiments}
Numerical experiments are carried out in order to investigate the influence of the $\beta$-VAE regularization on the characteristics of the latent space as well as the applicability to small data sets. 
For this purpose, objective error metrics are defined in Section~4.1 in terms of the spatial two-point correlation.
Then, using a simple synthetic data set, the influence of~$\beta$ on the reconstruction quality as well as the structure of the data set is investigated in Section~4.2.
Finally, based on various test cases with real and synthetic materials, the applicability to extremely small data sets is tested in Section~4.3.

For this purpose, two different model architectures are chosen in Section 4.2 and 4.3, respectively.
With the number of parameters in each layer being specified in the corresponding sections, the remaining hyperparameters are given in Table~1.
The numerical experiments are carried out on a test bench equipped with an AMD Ryzen 9 5900X CPU (overclocked to 4.8~GHz), 64~GB memory and an RTX 3080Ti GPU (overclocked to 1900~MHz).
Fully training the hybrid model takes around 2~hours for Section~4.2 and 30~minutes for Section~4.3, respectively.
However, the current implementation of the differentiable augmentations requires intense communication between GPU and CPU, and thus training speed might differ for hardware with different latency and bandwidth.
\begin{table}[b]
    \centering
    \begin{tabular}{l | c c l}
        \toprule
        Parameter & Section 4.2 & Section 4.3 & Description \\
        \midrule
Sample size & $6 \cdot 10^4$ & $16$ & \\
Batch size & $32$ & $4$ & \\
$\text{z}_{\text{dim}}$ & $5$ & $32$ & Dimensions of the latent variable\\
Optimizer & Adam & Adam & \\
$\alpha_{\text{gen}}$ & $10^{-4}$ & $10^{-4}$ & Learning rate of the mutual generator\\
$\alpha_{\text{disc}}$ & $4 \cdot 10^{-5}$ & $4 \cdot 10^{-5}$ & Learning rate of the discriminator\\
$\alpha_{\text{enc}}$ & $10^{-4}$ & $10^{-4}$ & Learning rate of the encoder\\
$\beta$ & $0.01 \leq \beta \leq 50$ & $1$ & Encoder loss multiplier for KL-loss\\ 
Epochs & $200$ & $2 \cdot 10^{4}$ & Total number of training epochs\\
$\lambda_{\text{vae}}$ & $1$ & $1$ & GAN loss multiplier for generations from $z$\\
$\lambda_{\text{noise}}$ & $1$ & $1$ & GAN loss multiplier for random generations\\
$\lambda_{\text{rec}}$ & $10^{-4}$ & $10^{-4}$ & GAN loss multiplier for reconstructions\\
        \bottomrule
    \end{tabular}
    \caption{Hyperparameters of the proposed ML model for the numerical experiments}
    \label{tab:hyperparameters}
\end{table}

\subsection{Error measures}
\label{sec:errormeasures}
Assessing the outcome of the model is one of the most important aspects in ML experiments to verify assumptions and measure performance.
In contrast to evaluating the loss value and the subjective reconstruction quality, using statistical descriptors as error metric provides a more objective and quantitative perspective as supplement to confirming visual similarities of generated microstructures. 
Specifically, the spatial two-point correlation is chosen to define a suitable error metric.
With applications dating back over 20 years~\cite{yeong_reconstructing_1998,torquato2002}, they are a well-established descriptor in the reconstruction community~\cite{bostanabad_computational_2018,seibert_microstructure_2022}.
A comprehensive introduction is given in~\cite{jiao_modeling_2007} and briefly summarized in the following.

Every sample of the microstructure can be referred as a realization of a specific random process, and the collection of all possible realizations is an ensemble~\cite{torquato2002}.
For any multi-phase random medium, given an indicator function $\mathcal{I}^{(i)}(\vec x^n)$, which has the value $1$ if the point\footnote{Following the common notation in machine learning, an image, e.g. a micrograph, is denoted as~$x$, since it is the natural representation of the input data as well as the training data. In contrast,~$\vec x$ is used to denote a spatial position.} $\vec x^n$ lies in phase $i$ and $0$ otherwise, the $n$-point autocorrelation of phase~$i$ is defined as 
\begin{equation}
   S_{n}^{(i)}(\vec x^1, \vec x^2, \dots, \vec x^n) = \underset{m \to \infty}{\mathrm{lim}} \left\langle {\mathcal{I}^{(i)}(\vec x^1)\mathcal{I}^{(i)}(\vec x^2) \dots \mathcal{I}^{(i)}(\vec x^n)} \right\rangle^m \; , 
\end{equation}
where $\langle \dotsc \rangle^m$ denotes the ensemble average over $m$ realizations. 
The equation can be further interpreted as the probability~$\mathcal{P}$ of $n$ points at chosen positions $\vec x^1, \vec x^2, \dots, \vec x^n$ are in phase $i$
\begin{equation}
   S_{n}^{(i)}(\vec x^1, \vec x^2, \dots, \vec x^n) = \mathcal{P} \left\{\mathcal{I}^{(i)}(\vec x^1), \mathcal{I}^{(i)}(\vec x^2), \dots,\mathcal{I}^{(i)}(\vec x^n)\right\} \; .
\end{equation} 

The two-point autocorrelation $S_{2}^{(1)}(\vec x^{1}, \vec x^{2})$ follows from this as a special case with~$n=2$ and constitutes a commonly chosen trade-off between information content (higher $n$) and computational efficiency (lower $n$).
Moreover, in a random heterogeneous medium,~$S_2$ does not depend on the absolute positions~$\vec x^1$ and~$\vec x^2$, but only on their relative displacement~$\vec r=\vec x^2 - \vec x^1$.
Hence, in the following, the spatial autocorrelation is expressed as~$S_{2}^{(1)}(\vec r)$.

It is worth noting that the volume fraction~$v_f$ of a phase is given as the autocorrelation with the zero-vector~$v_f = S_{2}^{(1)}(\vec 0)$.

In order to symmetrically quantify the difference between original and reconstructed microstructures, the root-mean-square error of the differences between the correlation functions~$S_{2,\text{A}}$ and~$S_{2,\text{B}}$ of microstructures~A and~B is defined as
\begin{equation} 
    \mathcal{E}_\text{A,B} = \sqrt{\dfrac{\bigintsss \left[S_{2,\text{A}}^{(1)}(\vec r)-S_{2,\text{B}}^{(1)}(\vec r)\right]^2 \text{d} \vec r }{ \int \text{d} \vec r }} \; .
\end{equation}
For the reconstruction error~$\mathcal{E}_\text{rec}$ of given microstructures from the corresponding latent space, structures~A and~B are simply the original and reconstructed structure, respectively.
In contrast, for the generation from random latent points, the error~$\mathcal{E}_\text{gen}$ of a realization is defined as the error with respect to the closest point in the training data set.
Both,~$\mathcal{E}_\text{rec}$ as well as~$\mathcal{E}_\text{gen}$ are averaged over all realizations for a given material.
Each structure is described by integer-valued indicator functions that are obtained by element-wise rounding.
The computation of the descriptors is carried out using the open-source software~\emph{MCRpy}~\cite{seibert_microstructure_2022}.

With this error measure as a tool for quantitative quality assessment, different numerical experiments are carried out in Sections~4.2 and~4.3. 

\subsection{Interpretability of latent space}
\label{sec:numericalexperiments_latentspace}
The focus of this section is to investigate the interpretability of the latent space.
For this purpose, a training data set is created comprising $60{,}000$ computationally generated microstructures with a resolution of $32 \times 32$ pixels.
Each structure contains a single elliptical inclusion.
The area fraction of the inclusions are set to be greater than $1.2 \%$ to maintain their visibility, and smaller than $19 \%$ to avoid any oversized inclusions.
Furthermore, the aspect ratios of the ellipses' semi-axes~$a$ and~$b$ are limited to~$1 \leq a/b \leq 4$, so that the ellipses are not overly stretched. 
In order to eliminate the influence of how boundaries are treated, inclusions are only placed such that they lie entirely within the image and are not cut by the boundary.
This allows any sample in the data set to be described by merely $5$ parameters, namely the lengths of the semi-major and -minor axes $a$ and $b$, the coordinates $\vec x = [x_1, \, x_2]$ of the center of the ellipse and the angle $\varphi$ between its major axis and the horizontal axis of the coordinate system.
Since five is the intrinsic dimensionality of the data set, the encoder is expected to find a latent representation of five variables that can adequately represent the data.

The model given in Table~2 is trained with various values for the regularization parameter $\beta$ ranging from $0.01$ to $50$, and all trained models are configured to perform reconstructions from the same reference sample.
The reference sample contains an ellipse, whose principal axes have the length of $6$ and $14$ pixels in the horizontal and vertical direction, respectively. 
The inclusion is located at the center of the microstructure with an angle of~$\varphi = \pi / 2$.
\begin{table}[hb]
    \centering
    \caption{Model architecture for numerical experiments of elliptical inclusions.}
    \begin{tabular}{r|c c c c c c c c}
        \toprule
& Layer & Type & Filter & Kernel & Strides & Padding & Batch Norm. & Activation \\
        \midrule
        \multirow{4}{1em}{\begin{turn}{90} $\;$ Enc. \end{turn}} &
        1 & Conv2D & 16 & $4 \times 4$ & 2 & Same & No & Leaky ReLU \\
        & 2 & Conv2D & 32 & $4 \times 4$ & 2 & Same & Yes & Leaky ReLU \\
        & 3 & Conv2D & 64 & $4 \times 4$ & 2 & Same & Yes & Leaky ReLU\\
        & 4 & Conv2D & 10 & $4 \times 4$ & 1 & Valid & Yes & Leaky ReLU\\
        \midrule
        \multirow{4}{1em}{\begin{turn}{90} $\;$ Gen. \end{turn}} &
        1 & TranspConv2D & 64 & $4 \times 4$ & 1 & Valid & Yes & Leaky ReLU\\
        & 2 & TranspConv2D & 32 & $4 \times 4$ & 2 & Same & Yes & Leaky ReLU\\
        & 3 & TranspConv2D & 16 & $4 \times 4$ & 2 & Same & Yes & Leaky ReLU\\
        & 4 & TranspConv2D & 1 & $4 \times 4$ & 2 & Same & Yes & Sigmoid\\
        \midrule
        \multirow{4}{1em}{\begin{turn}{90} $\;$ Disc. \end{turn}} &
        1 & Conv2D & 8 & $4 \times 4$ & 2 & Same & No & Leaky ReLU\\
        & 2 & Conv2D & 16 & $4 \times 4$ & 2 & Same & Yes & Leaky ReLU\\
        & 3 & Conv2D & 32 & $4 \times 4$ & 2 & Same & Yes & Leaky ReLU\\
        & 4 & Conv2D & 1 & $4 \times 4$ & 1 & Valid & Yes & Sigmoid\\
        \bottomrule
    \end{tabular}
    \label{tab: structure: model*}
\end{table}

Figure 3 presents the relation between the KL loss multiplier $\beta$ and the reconstruction error $\mathcal{E}_\text{rec}$.
It can be seen that for small~$\beta$, the descriptor error is close to~$0$.
Furthermore, it can be observed that the model produces worse reconstructions with increasing weights on the regularization error in the encoder loss function, although the reconstruction error has a local minimum around $\beta = 35$.
This result is plausible, since the minimization of the generator loss is a multi-objective optimization, and a higher weight on the regularization term decreases the priority of the reconstruction quality.
\begin{figure}
    \centering
    \includegraphics[scale=0.6]{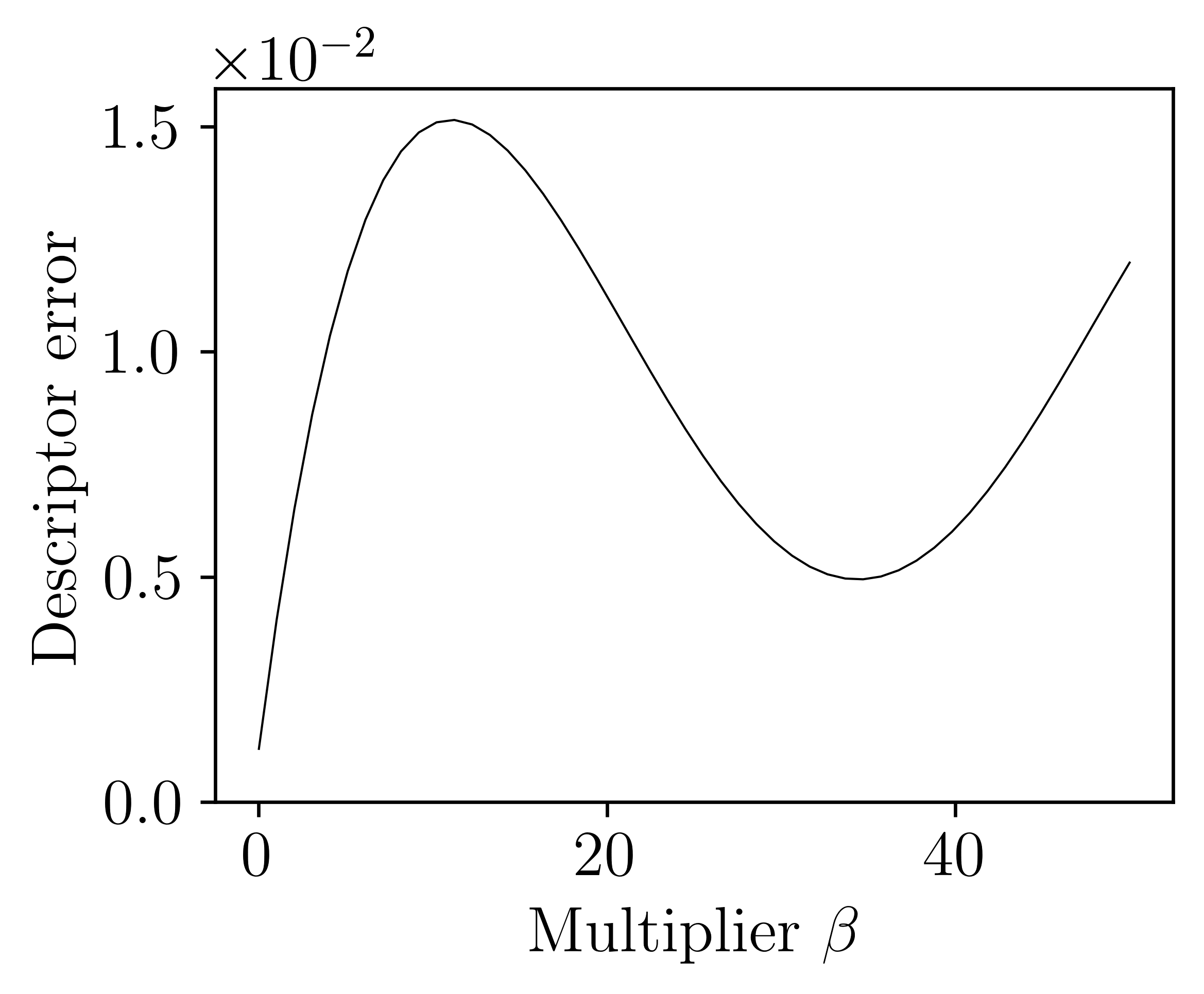}
    \caption{Descriptor error~(12) for the reconstruction of microstructures with elliptical inclusions as a function of the regularization parameter~$\beta$ in the encoder loss.}
    \label{fig:my_label} 
\end{figure}

Given a latent vector $z$ encoded from the reference sample, the generator should produce the exact reconstruction of the sample.
Assuming the latent space distribution matches a normal Gaussian distribution with $\sigma = 1$ and $\mu = 0$, so-called traversal plots are created by varying one latent variable of the latent vector $z$ at a time within a fixed interval of $\pm \, 3 \, \sigma$ to investigate the impact of each individual latent variable. 
Taking $\beta = 1$ as example, the traversal plot is displayed in Figure~4~(b). 
\begin{figure}[htpb!]
    \centering
    \subfigure[Traversal plot with $\beta = 0.01$]{\includegraphics[scale=0.46]{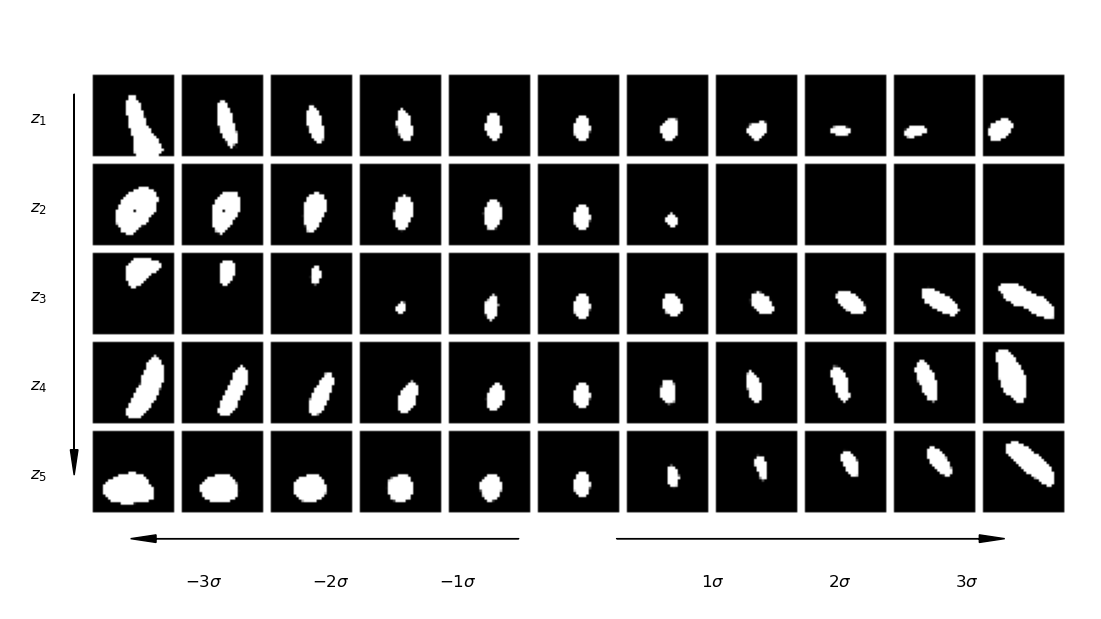}}
    \subfigure[Traversal plot with $\beta = 1$]{\includegraphics[scale=0.46]{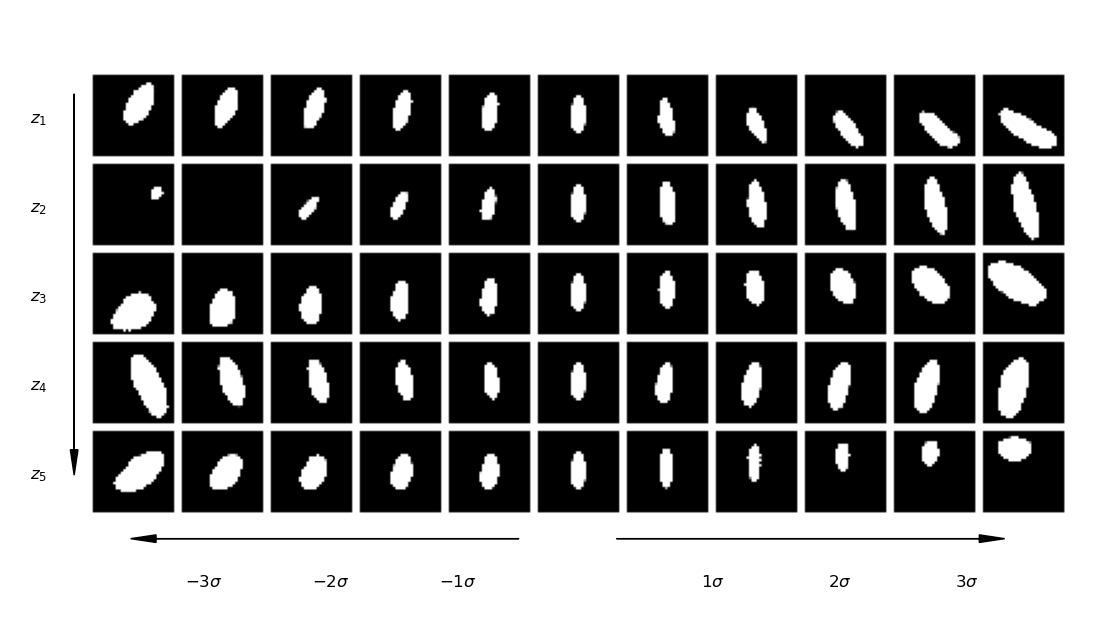}}
    \subfigure[Traversal plot with $\beta = 50$]{\includegraphics[scale=0.46]{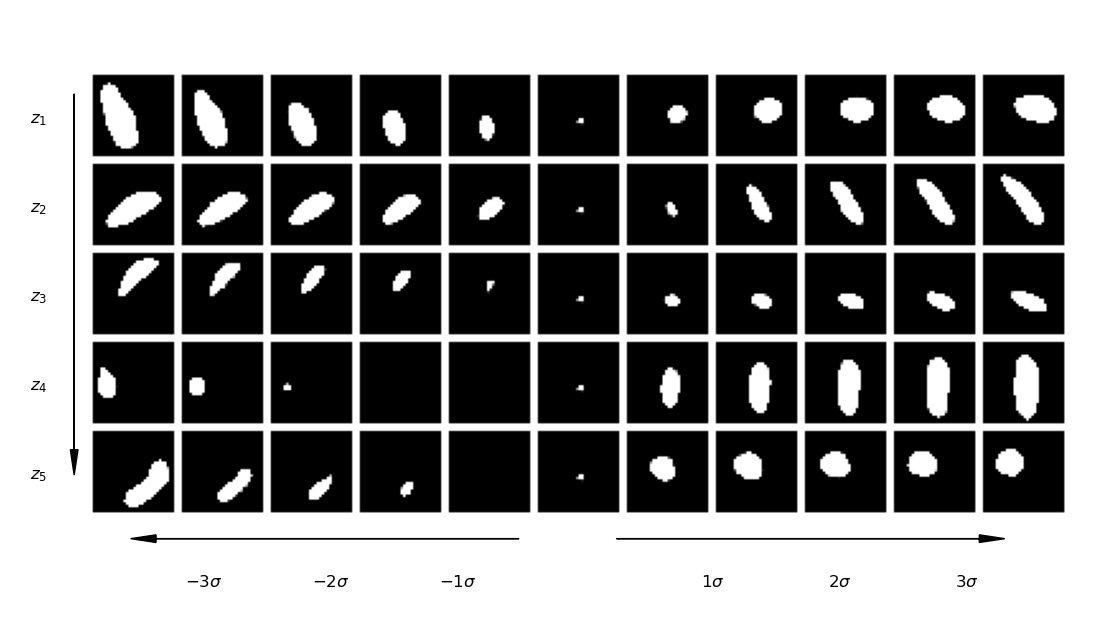}}
    \caption{Traversal plots with different values for the regularization parameter~$\beta$. Larger values increase the degree of order and interpretability of the latent space at the cost of decreased reconstruction quality. A feasible trade-off is achieved for~$\beta \approx 1$.} 
\end{figure}

As recognizable from the traversal plot, $z_1$ represents an anti-clockwise rotation, elongation and vertical translation with increased value.
Similarly, $z_2$ also rotates the inclusion while simultaneously increasing its volume.
The elongation, vertical displacement and volume are encoded in~$z_3$.
In contrast, $z_4$ is the only variable to clearly represent horizontal displacement.
Finally, $z_5$ affects the volume and horizontal extent of the inclusion for~$z_5<0$ and the vertical displacement and elongation for~$z_5>0$.
In summary, although the original degrees of freedom, i.e. coordinates, semi-axes and orientation, cannot be separated in the latent space, each coordinate can be associated with certain effects on the structure.
Most importantly, with the exception of very low values for~$z_2$, no sudden change can be observed between two consecutive images and the effect of increasing or decreasing a latent variable is relatively independent of its value.
In summary, the~$32^2$ parameters defining the microstructure data set in pixel space are successfully reduced to merely five latent dimensions.

A comparison of this model with a very weak and strong regularization is given in Figures~4~(a) and~(c), respectively, for an intuitive discussion of the influence of~$\beta$ on the latent space.
With very small values of~$\beta$, it becomes increasingly difficult to interpret the role of individual dimensions of the latent space, since the effect of a variable on the result is manifold and shows a stronger dependence on the value of the same variable. 
For example,~$z_3$ in Figure~4~(a) corresponds to a shrinkage and horizontal translation of the inclusion between~$-3\sigma$ and~$-2 \sigma$, then a sudden vertical translation until~$-1 \sigma$ and finally combined horizontal and vertical translation, rotation and elongation.
Moreover, with increasing distance in the latent variables, the generated samples become less and less ellipsoidal.
This can be seen with~$z_2$, where low values lead to holes in the inclusion and high values make the inclusion disappear altogether.
This reflects the fact that the latent space is not strongly regularized to obey a multivariate Gaussian distribution.
In contrast, with increasing~$\beta$, the reconstruction quality becomes very poor.
This can be seen in the center of Figure~4~(c), where the original ellipsoid cannot even be reconstructed accurately. 
Hence, the high-$\beta$ model is regarded as unsuitable for microstructure reconstruction and the corresponding traversal plot is merely given for completeness.
Within the interval $0.5 \leq \beta \leq 2$, a relatively high reconstruction quality and interpretability is observed, hence the natural value~$\beta = 1$ is chosen as a recommended value and for the remainder of this work.

In summary, the numerical experiments in this section demonstrate the influence of the regularization parameter~$\beta$ on the reconstruction quality as well as the interpretability of the latent space of the model.
In the conducted study, a choice in the order of magnitude of~$\beta \approx 1$ shows great potential in achieving a smooth and interpretable latent space without deteriorating the reconstruction quality.
Such prospective opens up possibilities of MCR in altering only certain properties of the microstructure to even construct previously non-existent microstructures with desired characteristics.

\subsection{Performance on extremely small data sets}
\label{sec:numericalexperiments_datasets}
A major challenge in the field of microstructure reconstruction is training with small data sets.
As motivated in the introduction, this becomes especially relevant when applying MCR to computational materials design, for which, naturally, very little data is available.
In this section, the proposed hybrid model is trained to reconstruct and generate various types of microstructures to investigate its versatility and performance on extremely small data sets.

For this purpose, a variety of microstructures is taken from the literature, specifically from the work of Li et al.~\cite{li_transfer_2018}.
This data includes 2D images of a metallic alloy, carbonate, ceramics, copolymer, rubber ($\text{PMMASiO}_2$), sandstone as well as a three-phase rubber composite.
Furthermore, the materials knowledge system PyMKS introduced by Brough et al.~\cite{brough_materials_2017} is used for generating three additional synthetic microstructures as training samples.
For each type of the 10 microstructures, only a \emph{single} image of resolution $256 \times 256$ pixels is available, which is sub-sampled to regions of size~$64 \times 64$ pixels to form a training data set of $16$ samples. 
We train a distinct model for each of the~$10$ different materials from scratch with the architecture\footnote{It is worth noting that although the training data set is much smaller than in Section~4.2, the larger input resolution requires a higher number of parameters in the models.} given in Table~3.
The fact that these models can still be trained robustly demonstrates the utility of the presented data augmentation scheme.
\begin{table}[hbp!]
    \centering
    \caption{Model architecture for numerical experiments in the small-data regime.}
    \begin{tabular}{r|c c c c c c c c}
        \toprule
& Layer & Type & Filter & Kernel & Strides & Padding & Batch Norma. & Activation \\
        \midrule
        \multirow{4}{1em}{\begin{turn}{90} Enc. $\;$ \end{turn}} &
        1 & Conv2D & 16 & $4 \times 4$ & 2 & Same & No & Leaky ReLU \\
        & 2 & Conv2D & 16 & $4 \times 4$ & 2 & Same & Yes & Leaky ReLU \\
        & 3 & Conv2D & 32 & $4 \times 4$ & 2 & Same & Yes & Leaky ReLU\\
        & 4 & Conv2D & 64 & $4 \times 4$ & 2 & Same & Yes & Leaky ReLU\\
        & 5 & Conv2D & 64 & $4 \times 4$ & 1 & Valid & Yes & Leaky ReLU\\
        \midrule
        \multirow{4}{1em}{\begin{turn}{90} Gen. $\;$\end{turn}} &
        1 & TranspConv2D & 64 & $4 \times 4$ & 1 & Valid & Yes & Leaky ReLU\\
        & 2 & TranspConv2D & 32 & $4 \times 4$ & 2 & Same & Yes & Leaky ReLU\\
        & 3 & TranspConv2D & 32 & $4 \times 4$ & 2 & Same & Yes & Leaky ReLU\\
        & 4 & TranspConv2D & 16 & $4 \times 4$ & 2 & Same & Yes & Leaky ReLU\\
        & 5 & TranspConv2D & 1 & $4 \times 4$ & 2 & Same & No & Sigmoid\\
        \midrule
        \multirow{4}{1em}{\begin{turn}{90} Disc. $\;$\end{turn}} &
        1 & Conv2D & 4 & $4 \times 4$ & 2 & Same & No & Leaky ReLU\\
        & 2 & Conv2D & 8 & $4 \times 4$ & 2 & Same & Yes & Leaky ReLU\\
        & 3 & Conv2D & 16 & $4 \times 4$ & 2 & Same & Yes & Leaky ReLU\\
        & 4 & Conv2D & 32 & $4 \times 4$ & 2 & Same & Yes & Leaky ReLU\\
        & 5 & Conv2D & 1 & $4 \times 4$ & 1 & Valid & Yes & Sigmoid\\
        \bottomrule
    \end{tabular}
    \label{tab: structure: model}
\end{table}

As compensation to the very small training data set, the number of epochs is increased to $20{,}000$ to make sure that the number of optimization steps is similar to the model in Section~4.2 with a conventional-size data set.
Interestingly, a common pattern of loss function developments is shared between training of different microstructures; the ones from alloy are displayed in Figure 5.
When observing loss functions of GANs, it is natural that a convergence of all loss functions to zero cannot be expected, since the scrutiny of the discriminator affects the evaluation of the generator and vice-versa.
Furthermore, the stochastic optimization as well as the adversarial training naturally lead to high noise.
Hence, only the stability of the mean value and the variance of the loss allow for an insight into a notion of convergence, although even these measures should not be over-interpreted.
While the GAN components converge already at around $6{,}000$ epochs, the encoder's total loss as well as KL loss stabilize after around $17{,}500$ epochs.
Generally, after~$20{,}000$ epochs, a stable state achieved throughout all considered materials.
\begin{figure}[ht]
    \centering
    \subfigure[Discriminator loss]{\includegraphics[scale=0.7]{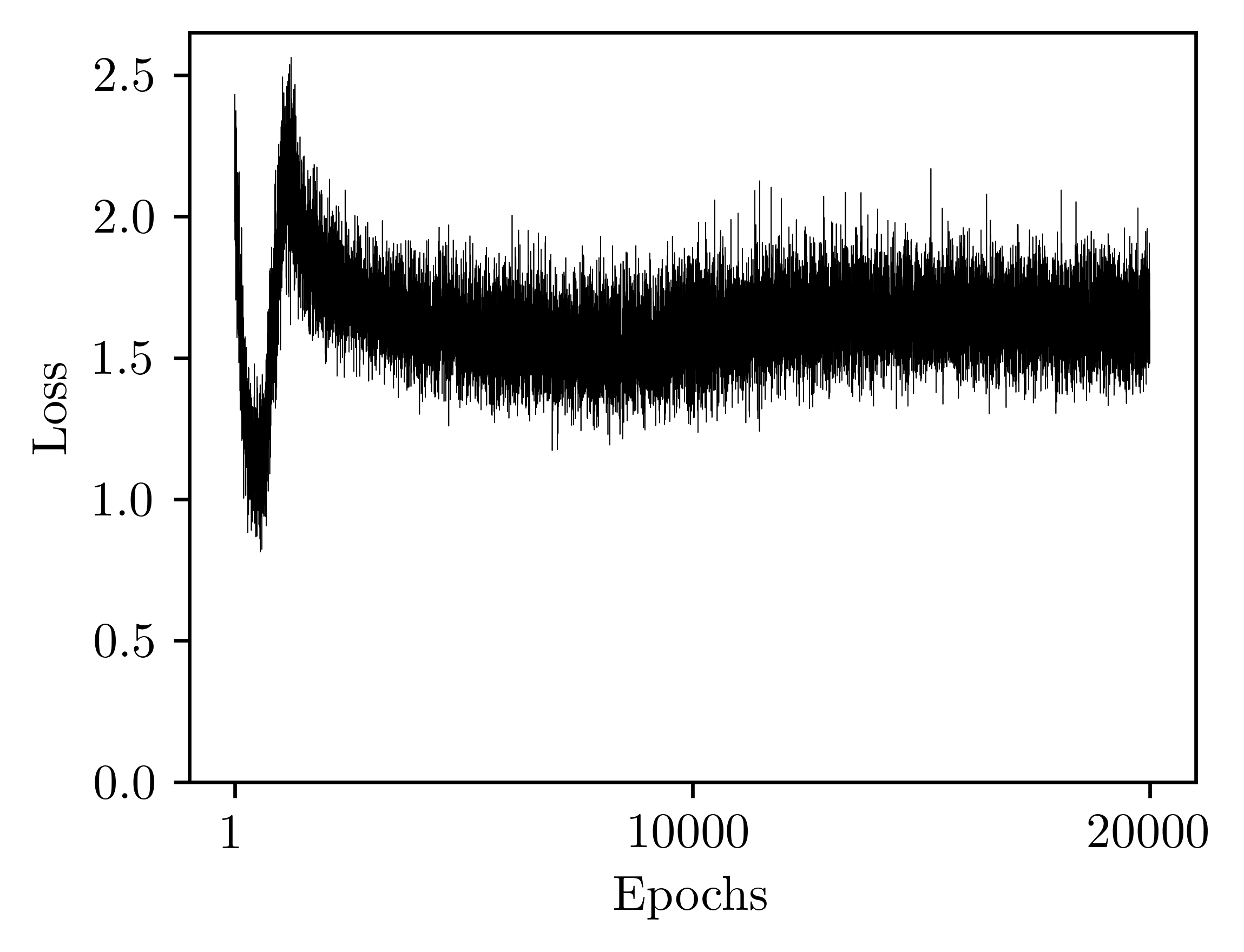}} 
    \subfigure[Generator loss]{\includegraphics[scale=0.7]{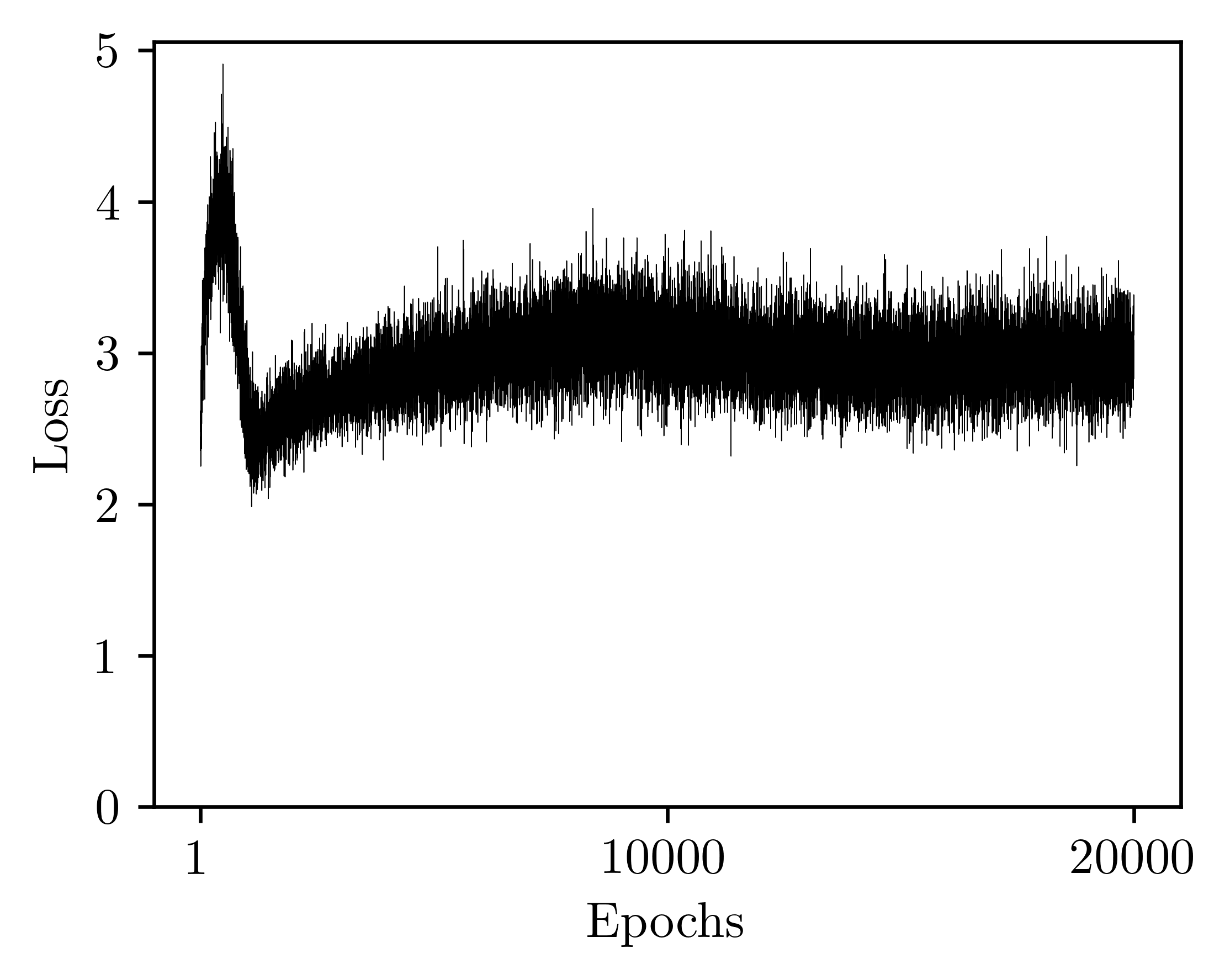}}
    \subfigure[Total encoder loss]{\includegraphics[scale=0.7]{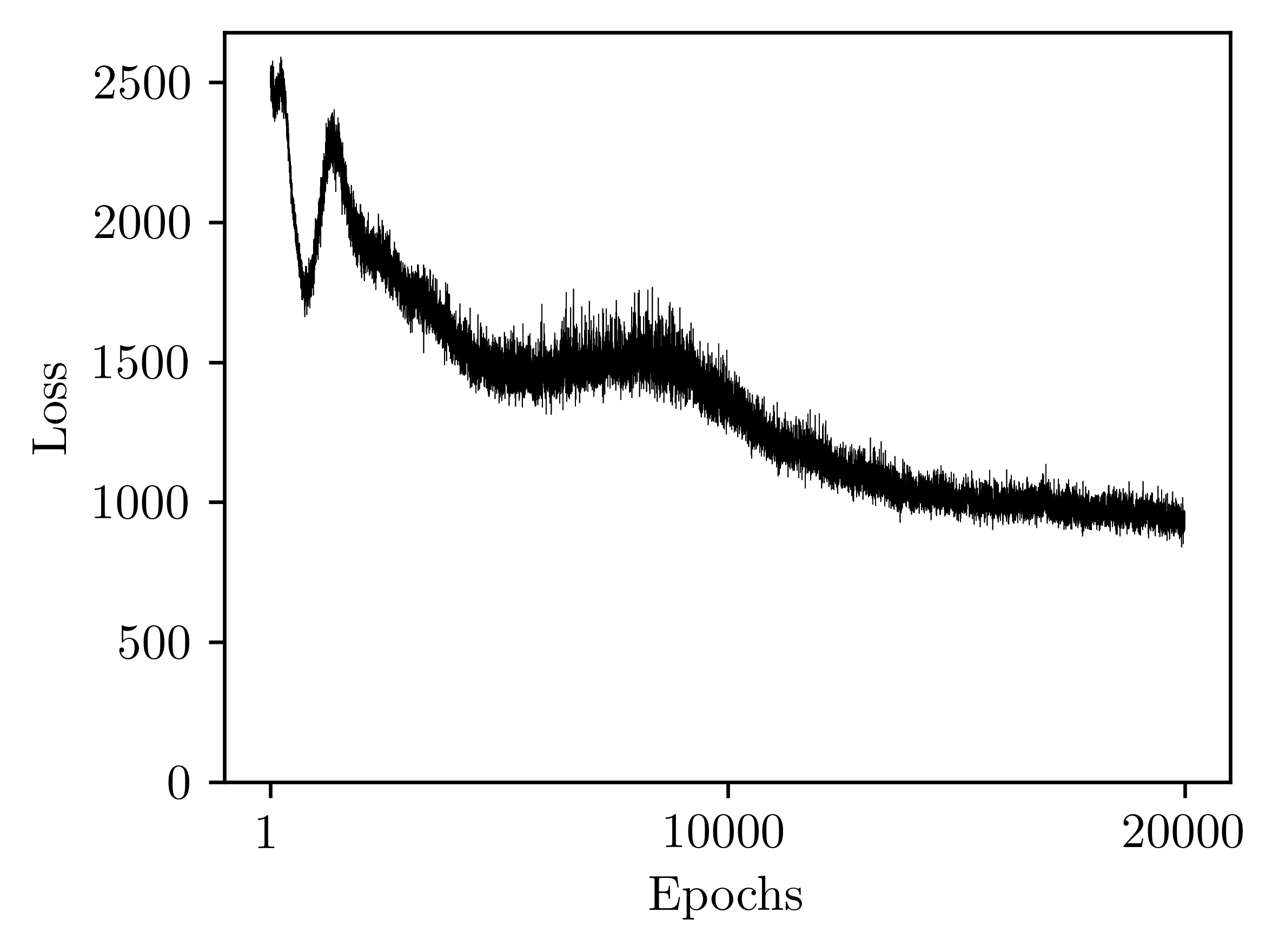}}
    \subfigure[Kullback-Leibler (KL) loss of the encoder]{\includegraphics[scale=0.7]{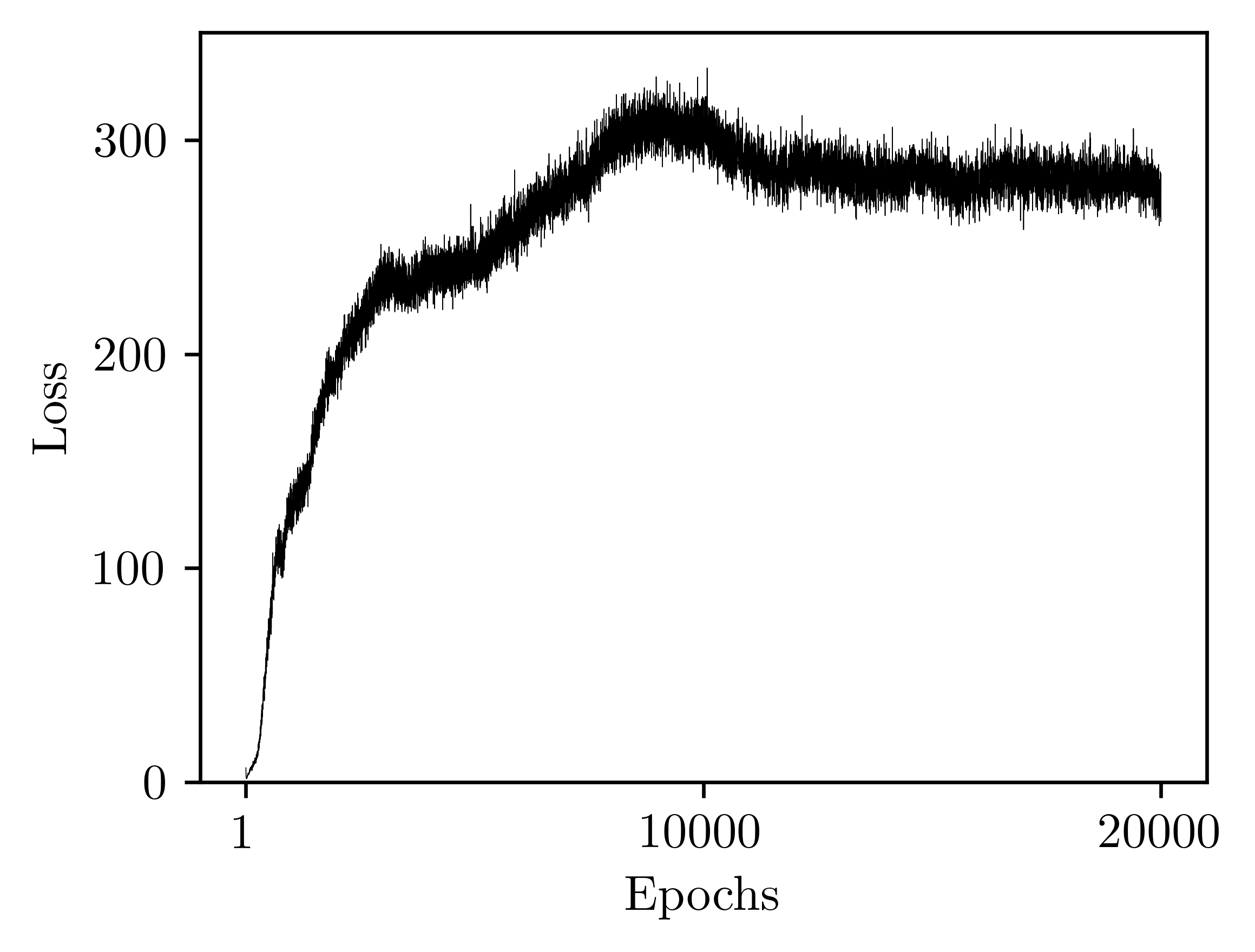}}
    \caption{Exemplary loss curves for training the proposed augmented hybrid model on the alloy data shown in Table~5. A similar behavior is observed in the loss curves for other materials (not shown here).}
\end{figure} 

\begin{table}[hbp!]
    \caption{Descriptor errors for generation~($\mathcal{E}_\mathrm{gen}$) and reconstruction~($\mathcal{E}_\mathrm{rec}$) of different microstructures reconstructed by the proposed augmented hybrid model.} 
    \centering
    \begin{tabular}{c | c  c} 
        \toprule
        Microstructure & $\mathcal{E}_\text{gen}$ & $\mathcal{E}_\text{rec}$ \\ 
        \midrule
        Alloy & $5.66 \cdot 10^{-3}$ & $4.66 \cdot 10^{-3}$\\
        Carbonate & $6.78 \cdot 10^{-3}$ & $8.16 \cdot 10^{-3}$ \\
        Ceramics & $1.71 \cdot 10^{-2}$ & $1.01 \cdot 10^{-2}$ \\
        Checkerboard & 0 & 0 \\
        Copolymer & $5.64 \cdot 10^{-2}$ & $6.96 \cdot 10^{-2}$ \\
        Rubber & $8.15 \cdot 10^{-4}$ & $3.08 \cdot 10^{-4}$ \\
        Synth. 1 & $1.06 \cdot 10^{-2}$ & $1.81 \cdot 10^{-2}$ \\
        Synth. 2 & $1.53 \cdot 10^{-2}$ & $4.53 \cdot 10^{-3}$ \\
        Synth. 3 & $1.89 \cdot 10^{-2}$ & $4.42 \cdot 10^{-3}$ \\
        Composite & $3.84 \cdot 10^{-2}$ & $9.71 \cdot 10^{-2}$ \\
        Sandstone & $5.97 \cdot 10^{-3}$ & $1.71 \cdot 10^{-2}$ \\
        \bottomrule
    \end{tabular}
    \label{tab:error}
\end{table}
\begin{figure}[hbp!]
    \centering
    \subfigure[Original]{\includegraphics[scale=0.3]{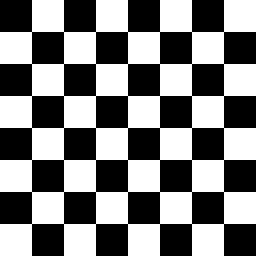}}
    \quad
    \hspace{1.5cm}
    \subfigure[Reconstruction]{\includegraphics[scale=0.3]{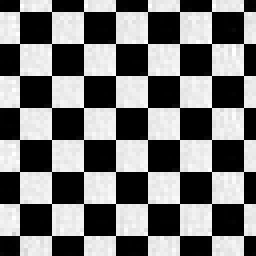}}
    \caption{Original and reconstructed checkerboard structure generated by the augmented hybrid model.}
\end{figure}

The introduced error metrics are listed in Table~4 for all microstructure types. 
The descriptor errors are mostly in the order of magnitude of~$1\%$ or lower, whereby generation and reconstruction lie in a similar order of magnitude.
The highest error is observed for the copolymer, for which spatial correlations are arguably not a very suitable descriptor, as demonstrated in~\cite{seibert_descriptor-based_2022,seibert_microstructure_2022}.
The lowest error is obtained for the checkerboard structure, where a perfect reconstruction is achieved as shown in Figure~6.
Despite noise at the white sections and a slight offset of the board, the checkerboard pattern is reconstructed perfectly with a clean separation between the bright and dark parts.
For this specific case, mode collapse of the generator is observed and expected, since the samples themselves do not exhibit any variations.
For the remaining materials, high-quality reconstruction results are achieved as can be seen in Table~5.
However, the model has the tendency to generate noisy images especially with larger continuous phases such as the ceramics.
A proper post-processing procedure as in~\cite{seibert_two-stage_2023} might be defined in the future to further improve the results.
Furthermore, it can be noted that despite the high difficulty of representing and reconstructing the grain boundary structure of the alloy, the model is able to capture these characteristics to some extent.
Even though a post-processing procedure would be required to bridge disconnected grain boundaries, the general trends and shapes are visually convincing.
Although the model is designed to focus on two-phase microstructures, the three-phase rubber composites is reconstructed well.
However, the third phase is mostly not constructed as a continuous phase when generated from a random latent vector. 
Finally, Table~6 shows more random generations, and no sign of mode collapse is observed for any type of microstructure.

\begin{table}[hbp!]
    \centering
    \begin{tabular}{r|c c|c c|c c c c}
    \toprule
& orig. & rec. & orig. & rec. & \multicolumn{4}{c}{generated from random $z$}\\
\midrule
\begin{turn}{90} $\;$ Alloy \end{turn} &
\includegraphics[width=0.095\linewidth]{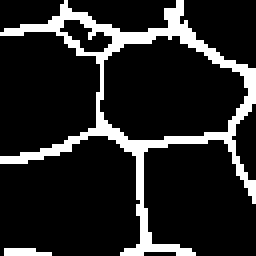} &
\includegraphics[width=0.095\linewidth]{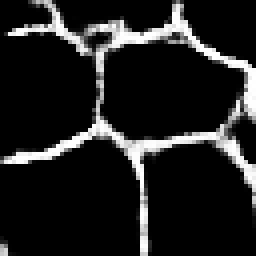} & 
\includegraphics[width=0.095\linewidth]{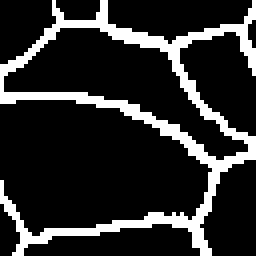} &
\includegraphics[width=0.095\linewidth]{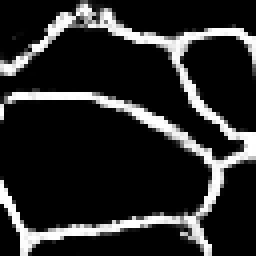} & 
\includegraphics[width=0.095\linewidth]{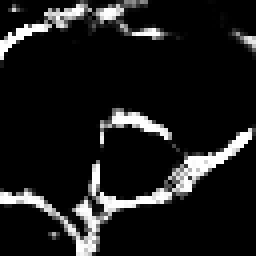} & 
\includegraphics[width=0.095\linewidth]{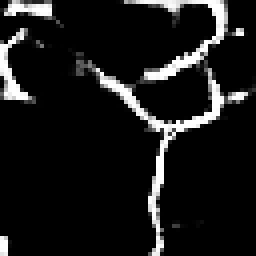} & 
\includegraphics[width=0.095\linewidth]{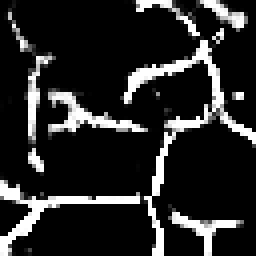} & 
\includegraphics[width=0.095\linewidth]{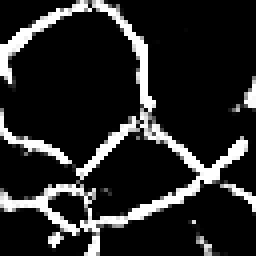}
\\
\begin{turn}{90} Carbon. \end{turn} &
\includegraphics[width=0.095\linewidth]{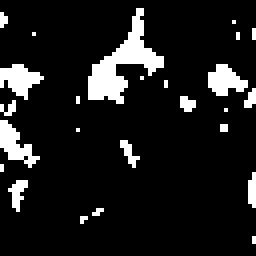} &
\includegraphics[width=0.095\linewidth]{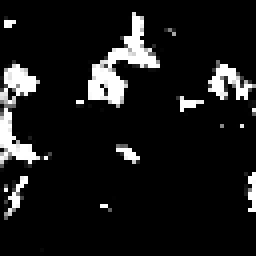} & 
\includegraphics[width=0.095\linewidth]{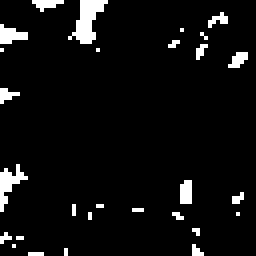} &
\includegraphics[width=0.095\linewidth]{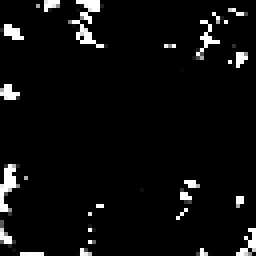} & 
\includegraphics[width=0.095\linewidth]{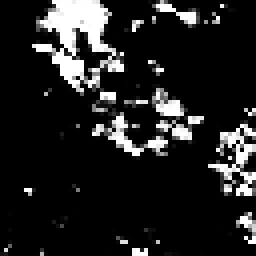} & 
\includegraphics[width=0.095\linewidth]{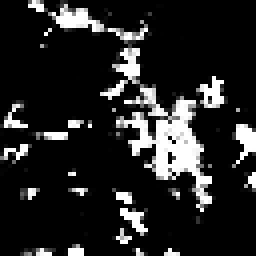} & 
\includegraphics[width=0.095\linewidth]{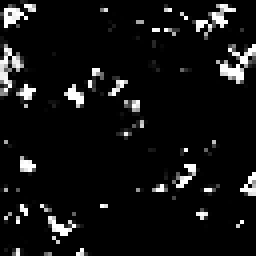} & 
\includegraphics[width=0.095\linewidth]{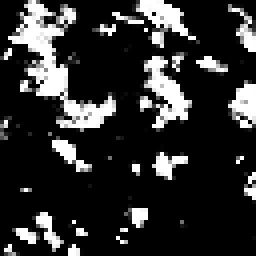}
\\
\begin{turn}{90} Ceramics \end{turn} &
\includegraphics[width=0.095\linewidth]{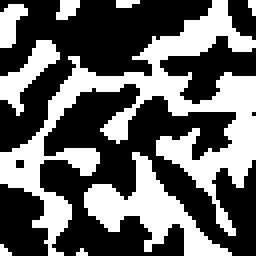} &
\includegraphics[width=0.095\linewidth]{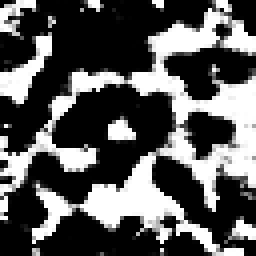} & 
\includegraphics[width=0.095\linewidth]{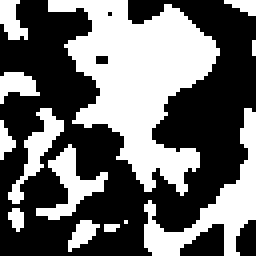} &
\includegraphics[width=0.095\linewidth]{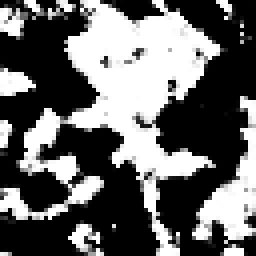} & 
\includegraphics[width=0.095\linewidth]{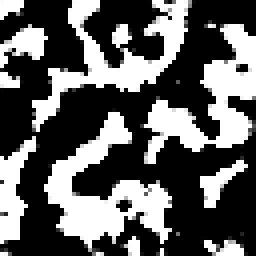} & 
\includegraphics[width=0.095\linewidth]{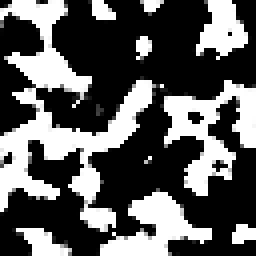} & 
\includegraphics[width=0.095\linewidth]{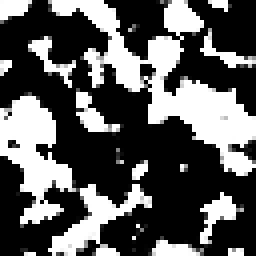} & 
\includegraphics[width=0.095\linewidth]{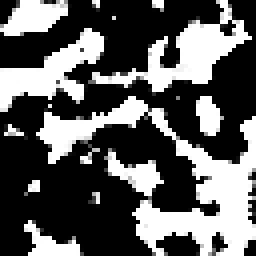}
\\
\begin{turn}{90} Copolym. \end{turn} &
\includegraphics[width=0.095\linewidth]{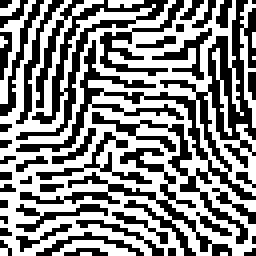} &
\includegraphics[width=0.095\linewidth]{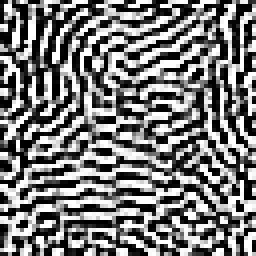} & 
\includegraphics[width=0.095\linewidth]{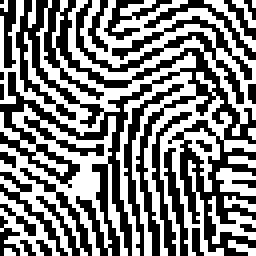} &
\includegraphics[width=0.095\linewidth]{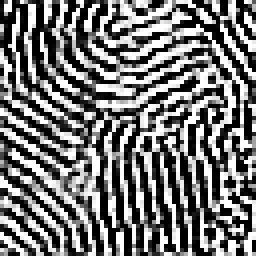} & 
\includegraphics[width=0.095\linewidth]{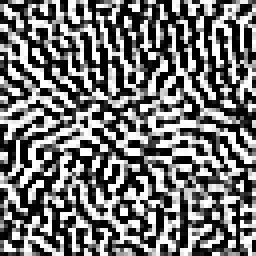} & 
\includegraphics[width=0.095\linewidth]{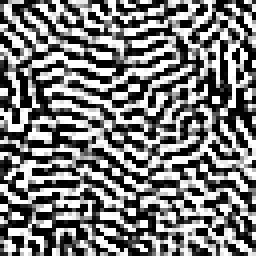} & 
\includegraphics[width=0.095\linewidth]{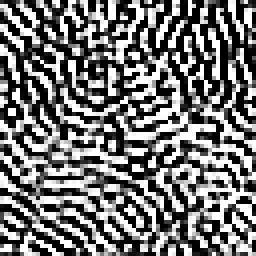} & 
\includegraphics[width=0.095\linewidth]{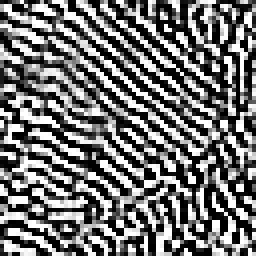}
\\
\begin{turn}{90} Rubber \end{turn} &
\includegraphics[width=0.095\linewidth]{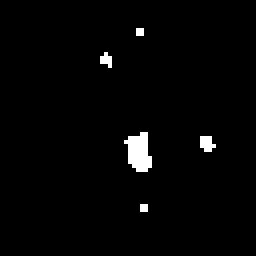} &
\includegraphics[width=0.095\linewidth]{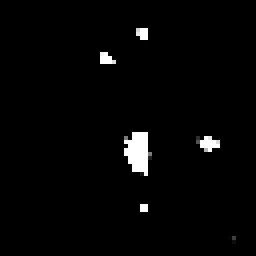} & 
\includegraphics[width=0.095\linewidth]{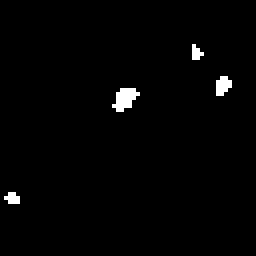} &
\includegraphics[width=0.095\linewidth]{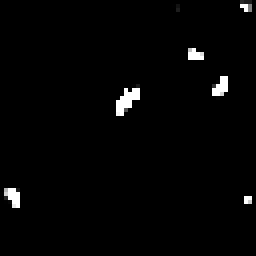} & 
\includegraphics[width=0.095\linewidth]{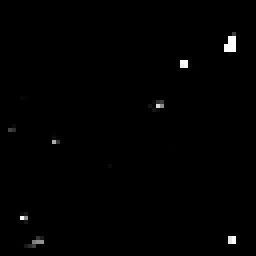} & 
\includegraphics[width=0.095\linewidth]{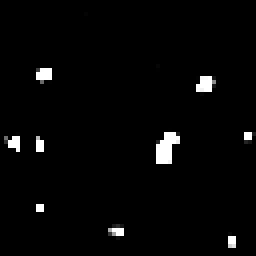} & 
\includegraphics[width=0.095\linewidth]{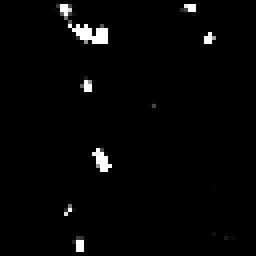} & 
\includegraphics[width=0.095\linewidth]{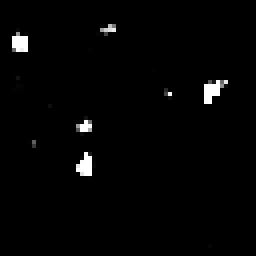}
\\
\begin{turn}{90} Synth. 1 \end{turn} &
\includegraphics[width=0.095\linewidth]{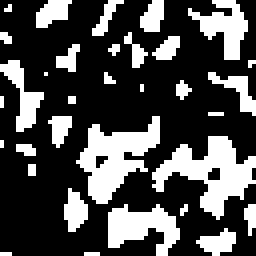} &
\includegraphics[width=0.095\linewidth]{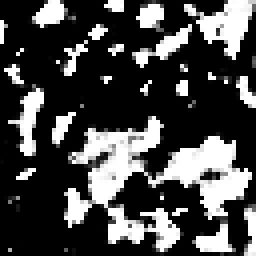} & 
\includegraphics[width=0.095\linewidth]{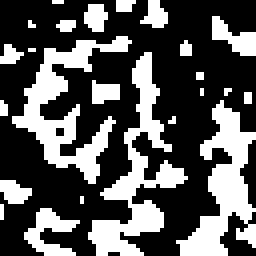} &
\includegraphics[width=0.095\linewidth]{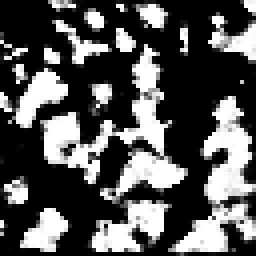} & 
\includegraphics[width=0.095\linewidth]{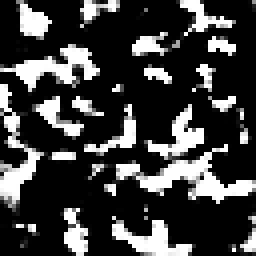} & 
\includegraphics[width=0.095\linewidth]{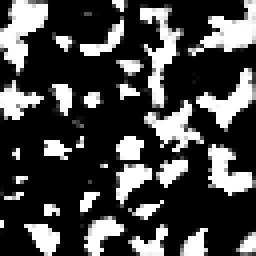} & 
\includegraphics[width=0.095\linewidth]{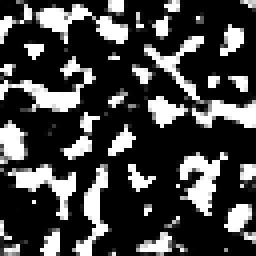} & 
\includegraphics[width=0.095\linewidth]{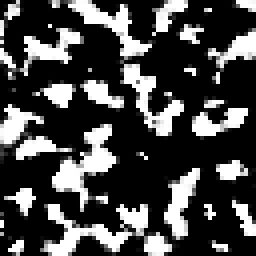}
\\
\begin{turn}{90} Synth. 2 \end{turn} &
\includegraphics[width=0.095\linewidth]{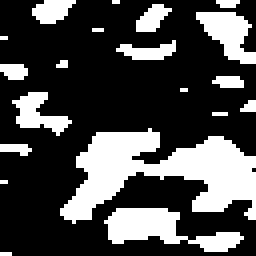} &
\includegraphics[width=0.095\linewidth]{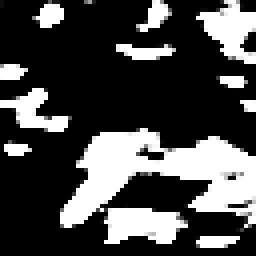} & 
\includegraphics[width=0.095\linewidth]{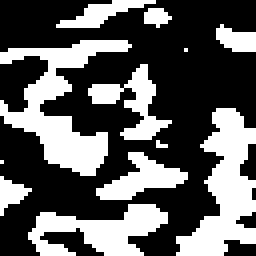} &
\includegraphics[width=0.095\linewidth]{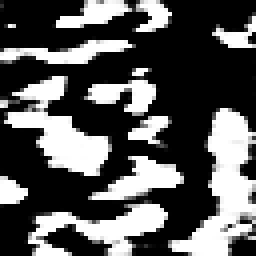} & 
\includegraphics[width=0.095\linewidth]{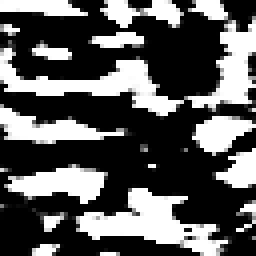} & 
\includegraphics[width=0.095\linewidth]{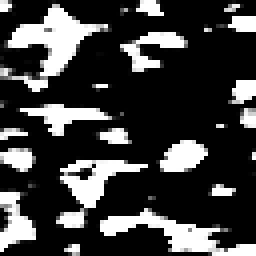} & 
\includegraphics[width=0.095\linewidth]{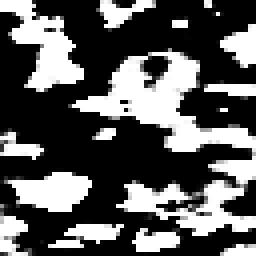} & 
\includegraphics[width=0.095\linewidth]{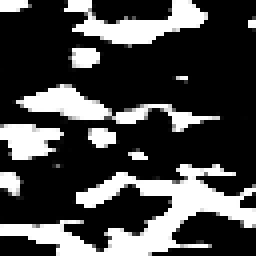}
\\
\begin{turn}{90} Synth. 3 \end{turn} &
\includegraphics[width=0.095\linewidth]{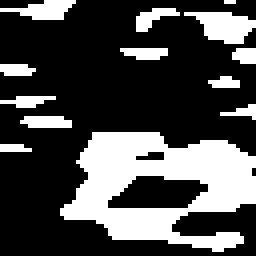} &
\includegraphics[width=0.095\linewidth]{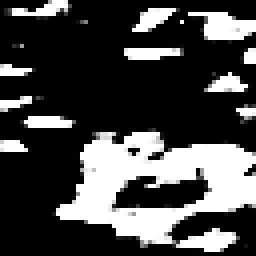} & 
\includegraphics[width=0.095\linewidth]{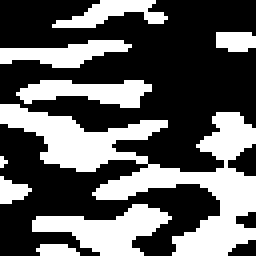} &
\includegraphics[width=0.095\linewidth]{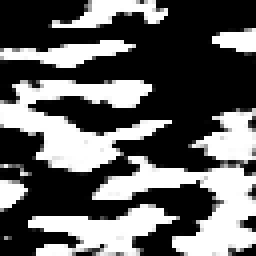} & 
\includegraphics[width=0.095\linewidth]{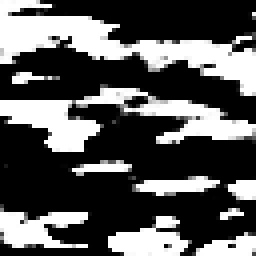} & 
\includegraphics[width=0.095\linewidth]{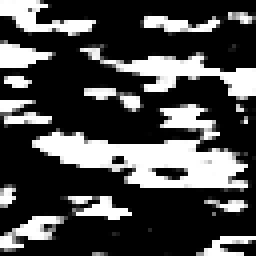} & 
\includegraphics[width=0.095\linewidth]{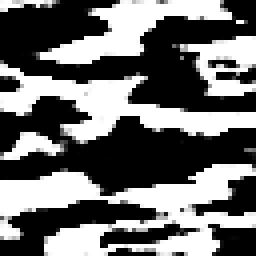} & 
\includegraphics[width=0.095\linewidth]{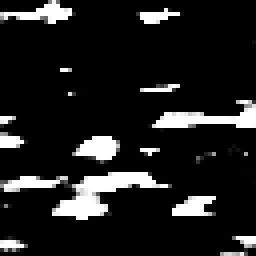}
\\
\begin{turn}{90} Compos. \end{turn} &
\includegraphics[width=0.095\linewidth]{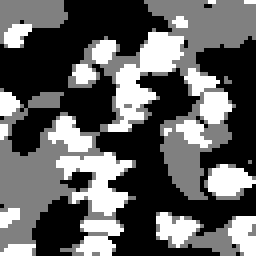} &
\includegraphics[width=0.095\linewidth]{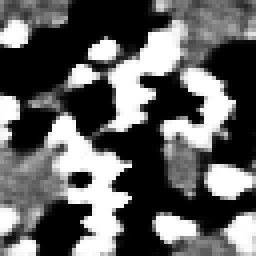} & 
\includegraphics[width=0.095\linewidth]{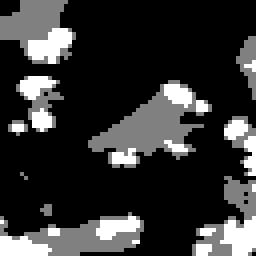} &
\includegraphics[width=0.095\linewidth]{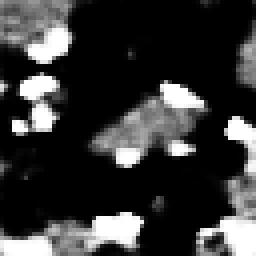} & 
\includegraphics[width=0.095\linewidth]{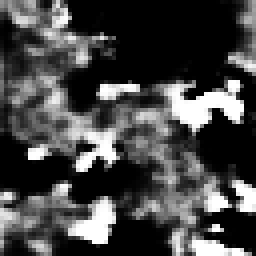} & 
\includegraphics[width=0.095\linewidth]{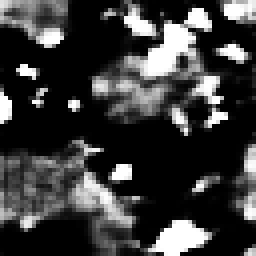} & 
\includegraphics[width=0.095\linewidth]{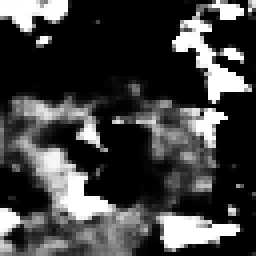} & 
\includegraphics[width=0.095\linewidth]{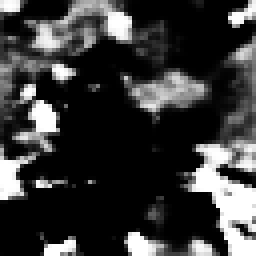}
\\
\begin{turn}{90} Sandst. \end{turn} &
\includegraphics[width=0.095\linewidth]{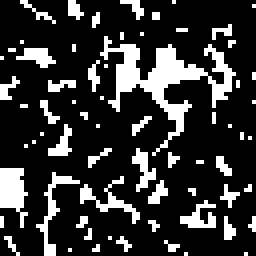} &
\includegraphics[width=0.095\linewidth]{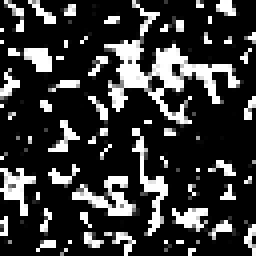} & 
\includegraphics[width=0.095\linewidth]{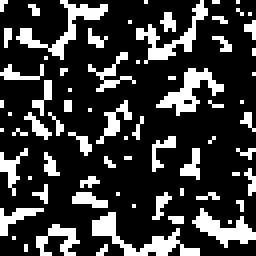} &
\includegraphics[width=0.095\linewidth]{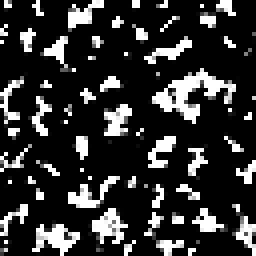} & 
\includegraphics[width=0.095\linewidth]{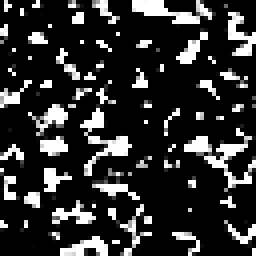} & 
\includegraphics[width=0.095\linewidth]{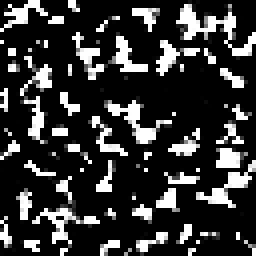} & 
\includegraphics[width=0.095\linewidth]{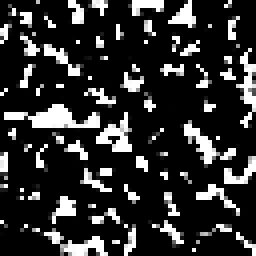} & 
\includegraphics[width=0.095\linewidth]{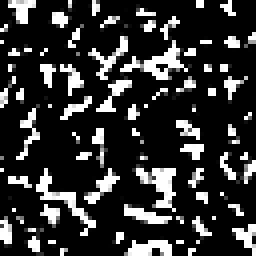}
\\
\bottomrule
    \end{tabular}
    \caption{Reconstructions from given structures (left) as well as synthetic structures generated from random points~$z$ in the latent space (right) using the trained proposed hybrid models.}
    \label{tab:my_label}
\end{table}

\begin{table}[hbp!]
    \centering
    \begin{tabular}{c|c  c  c  c  c  c  c  c  c}
    \toprule
\begin{turn}{90} $\;$ Alloy \end{turn} &
\includegraphics[width=0.09\linewidth]{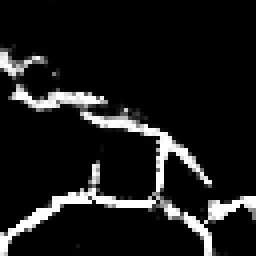} &
\includegraphics[width=0.09\linewidth]{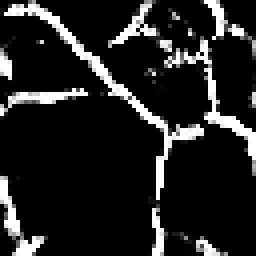} & 
\includegraphics[width=0.09\linewidth]{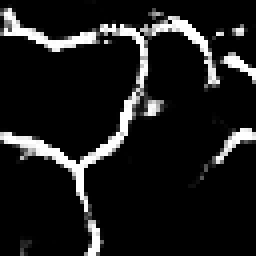} &
\includegraphics[width=0.09\linewidth]{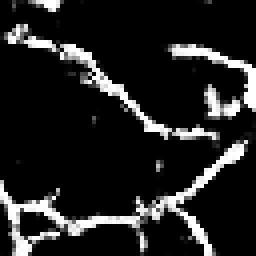} & 
\includegraphics[width=0.09\linewidth]{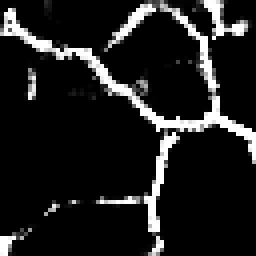} & 
\includegraphics[width=0.09\linewidth]{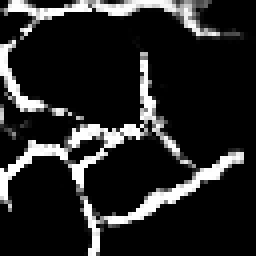} & 
\includegraphics[width=0.09\linewidth]{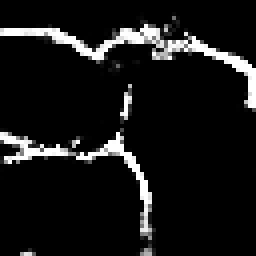} & 
\includegraphics[width=0.09\linewidth]{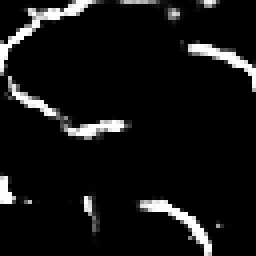} 
\\
\begin{turn}{90} Carbon. \end{turn} &
\includegraphics[width=0.09\linewidth]{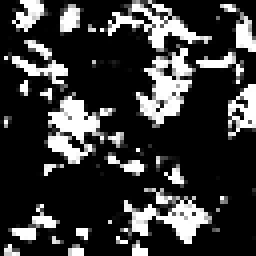} &
\includegraphics[width=0.09\linewidth]{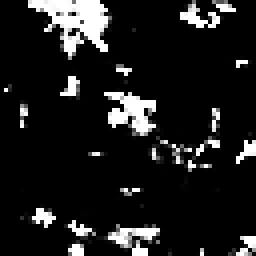} & 
\includegraphics[width=0.09\linewidth]{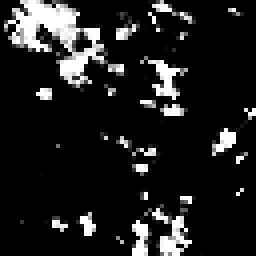} &
\includegraphics[width=0.09\linewidth]{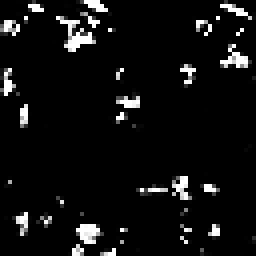} & 
\includegraphics[width=0.09\linewidth]{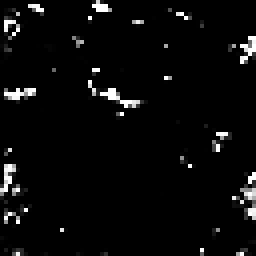} & 
\includegraphics[width=0.09\linewidth]{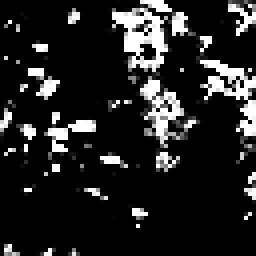} & 
\includegraphics[width=0.09\linewidth]{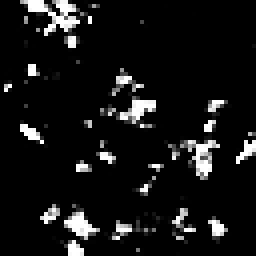} & 
\includegraphics[width=0.09\linewidth]{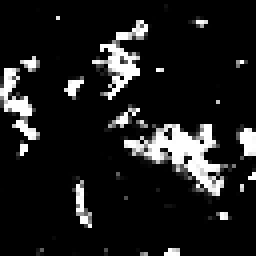} 
\\
\begin{turn}{90} Ceramics \end{turn} &
\includegraphics[width=0.09\linewidth]{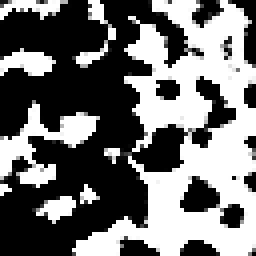} &
\includegraphics[width=0.09\linewidth]{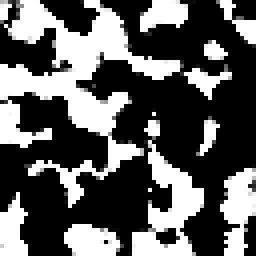} & 
\includegraphics[width=0.09\linewidth]{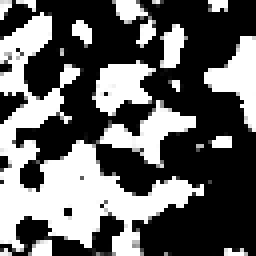} &
\includegraphics[width=0.09\linewidth]{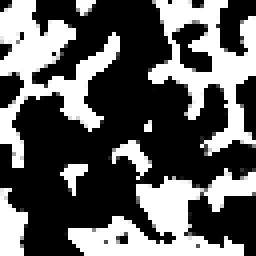} & 
\includegraphics[width=0.09\linewidth]{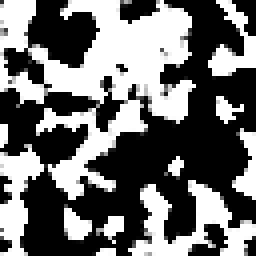} & 
\includegraphics[width=0.09\linewidth]{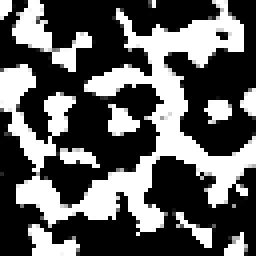} & 
\includegraphics[width=0.09\linewidth]{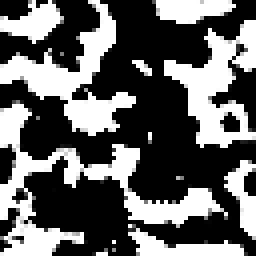} & 
\includegraphics[width=0.09\linewidth]{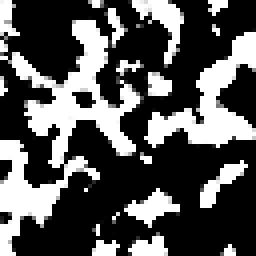} 
\\
\begin{turn}{90} Copolym. \end{turn} &
\includegraphics[width=0.09\linewidth]{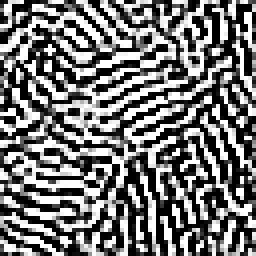} &
\includegraphics[width=0.09\linewidth]{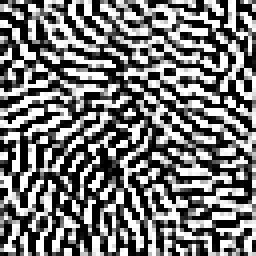} & 
\includegraphics[width=0.09\linewidth]{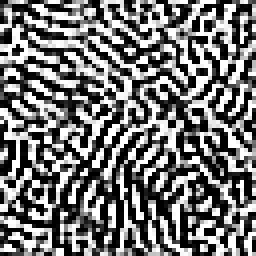} &
\includegraphics[width=0.09\linewidth]{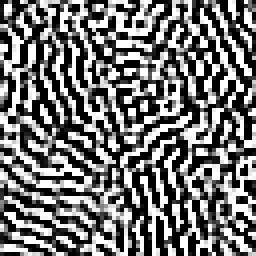} & 
\includegraphics[width=0.09\linewidth]{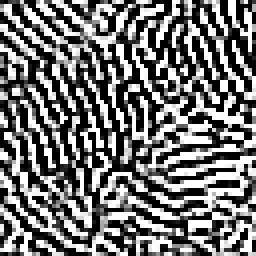} & 
\includegraphics[width=0.09\linewidth]{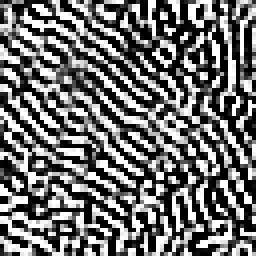} & 
\includegraphics[width=0.09\linewidth]{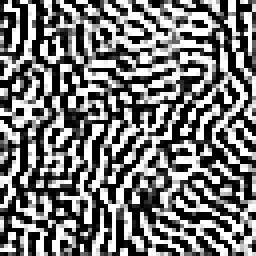} & 
\includegraphics[width=0.09\linewidth]{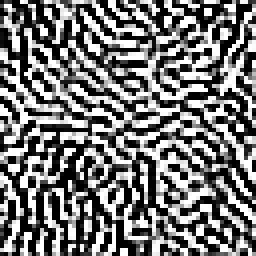} 
\\
\begin{turn}{90} Rubber \end{turn} &
\includegraphics[width=0.09\linewidth]{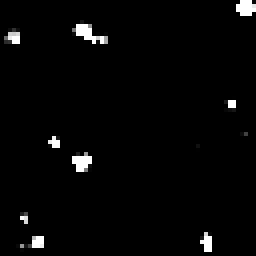} &
\includegraphics[width=0.09\linewidth]{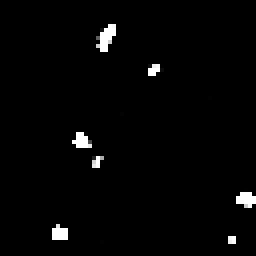} & 
\includegraphics[width=0.09\linewidth]{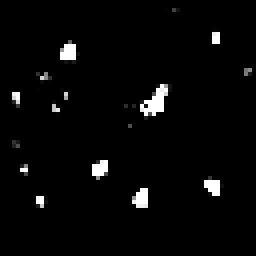} &
\includegraphics[width=0.09\linewidth]{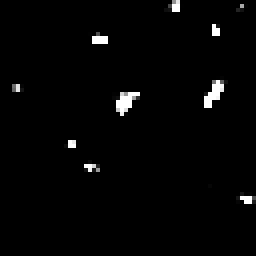} & 
\includegraphics[width=0.09\linewidth]{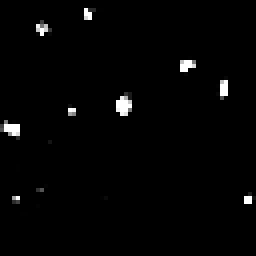} & 
\includegraphics[width=0.09\linewidth]{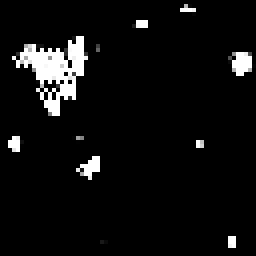} & 
\includegraphics[width=0.09\linewidth]{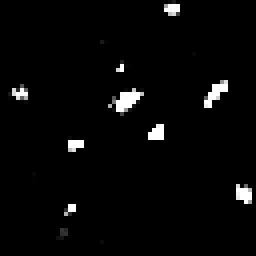} & 
\includegraphics[width=0.09\linewidth]{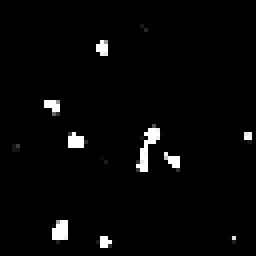} 
\\
\begin{turn}{90} Synth. 1 \end{turn} &
\includegraphics[width=0.09\linewidth]{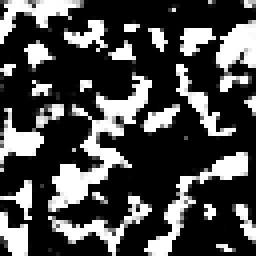} &
\includegraphics[width=0.09\linewidth]{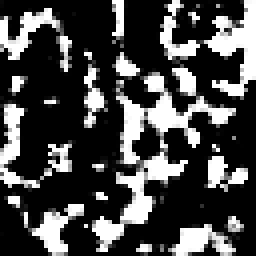} & 
\includegraphics[width=0.09\linewidth]{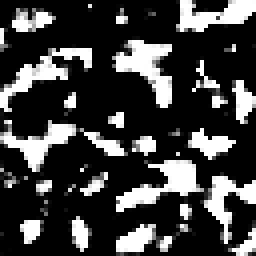} &
\includegraphics[width=0.09\linewidth]{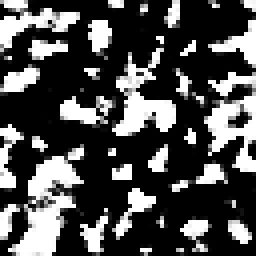} & 
\includegraphics[width=0.09\linewidth]{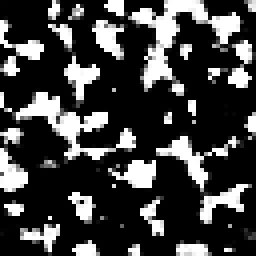} & 
\includegraphics[width=0.09\linewidth]{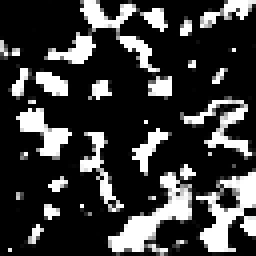} & 
\includegraphics[width=0.09\linewidth]{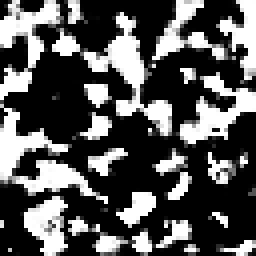} & 
\includegraphics[width=0.09\linewidth]{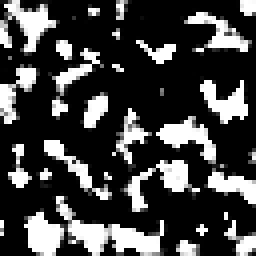} 
\\
\begin{turn}{90} Synth. 2 \end{turn} &
\includegraphics[width=0.09\linewidth]{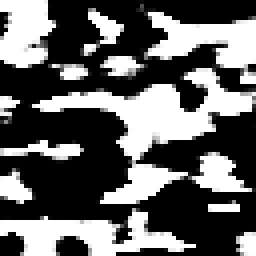} &
\includegraphics[width=0.09\linewidth]{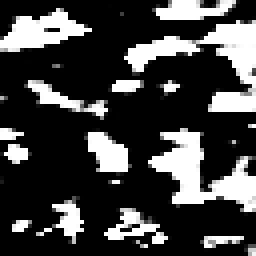} & 
\includegraphics[width=0.09\linewidth]{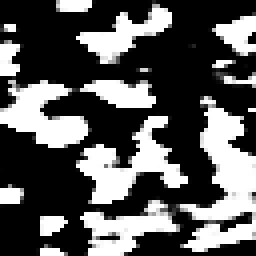} &
\includegraphics[width=0.09\linewidth]{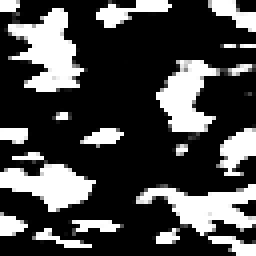} & 
\includegraphics[width=0.09\linewidth]{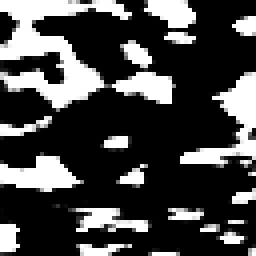} & 
\includegraphics[width=0.09\linewidth]{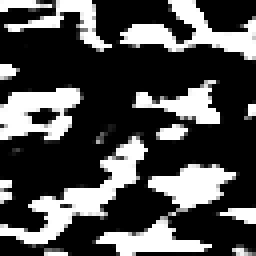} & 
\includegraphics[width=0.09\linewidth]{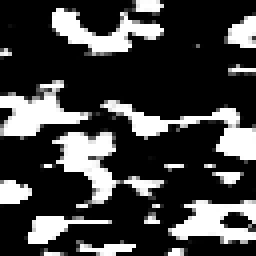} & 
\includegraphics[width=0.09\linewidth]{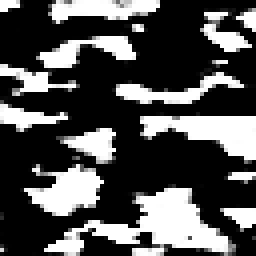} 
\\
\begin{turn}{90} Synth. 3 \end{turn} &
\includegraphics[width=0.09\linewidth]{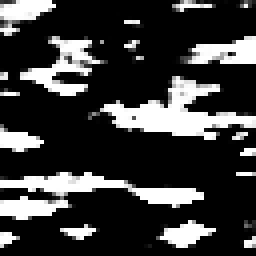} &
\includegraphics[width=0.09\linewidth]{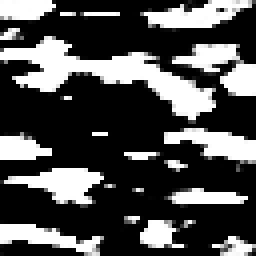} & 
\includegraphics[width=0.09\linewidth]{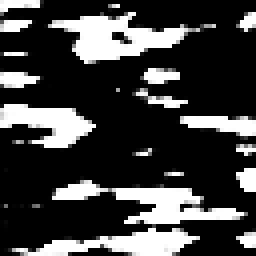} &
\includegraphics[width=0.09\linewidth]{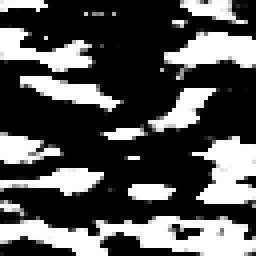} & 
\includegraphics[width=0.09\linewidth]{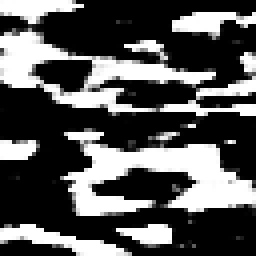} & 
\includegraphics[width=0.09\linewidth]{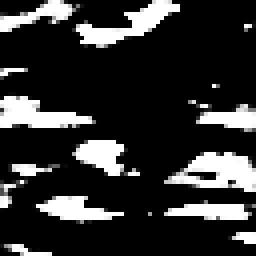} & 
\includegraphics[width=0.09\linewidth]{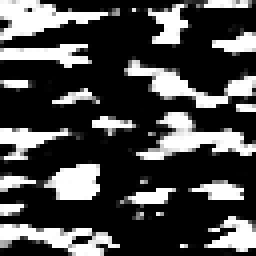} & 
\includegraphics[width=0.09\linewidth]{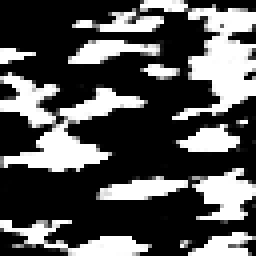} 
\\
\begin{turn}{90} Compos. \end{turn} &
\includegraphics[width=0.09\linewidth]{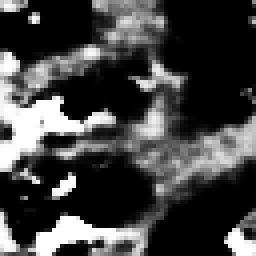} &
\includegraphics[width=0.09\linewidth]{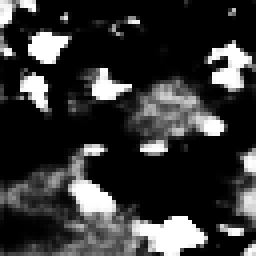} & 
\includegraphics[width=0.09\linewidth]{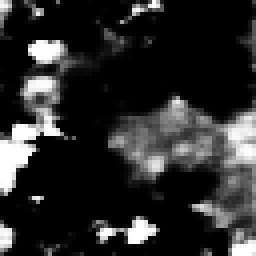} &
\includegraphics[width=0.09\linewidth]{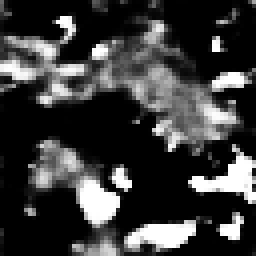} & 
\includegraphics[width=0.09\linewidth]{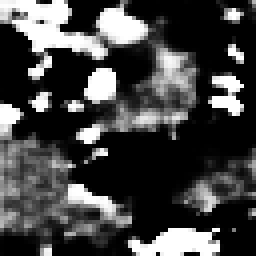} & 
\includegraphics[width=0.09\linewidth]{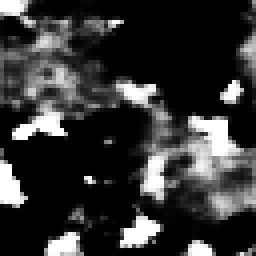} & 
\includegraphics[width=0.09\linewidth]{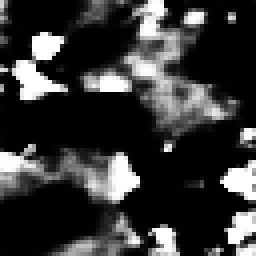} & 
\includegraphics[width=0.09\linewidth]{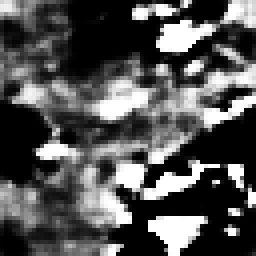} 
\\
\begin{turn}{90} Sandst. \end{turn} &
\includegraphics[width=0.09\linewidth]{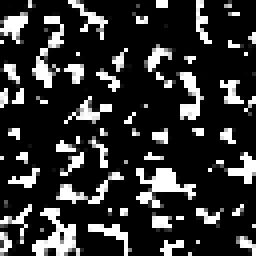} &
\includegraphics[width=0.09\linewidth]{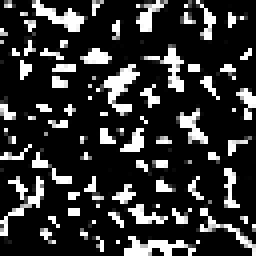} & 
\includegraphics[width=0.09\linewidth]{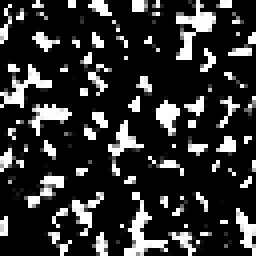} &
\includegraphics[width=0.09\linewidth]{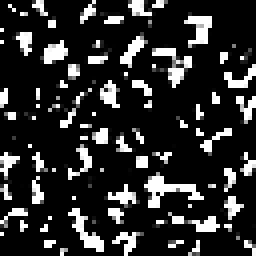} & 
\includegraphics[width=0.09\linewidth]{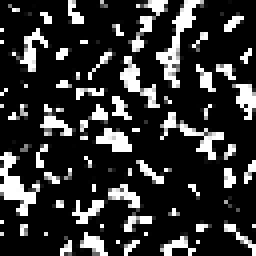} & 
\includegraphics[width=0.09\linewidth]{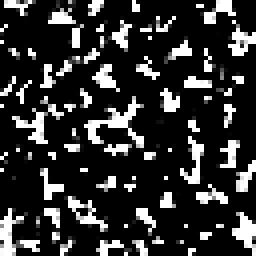} & 
\includegraphics[width=0.09\linewidth]{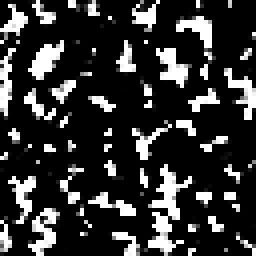} & 
\includegraphics[width=0.09\linewidth]{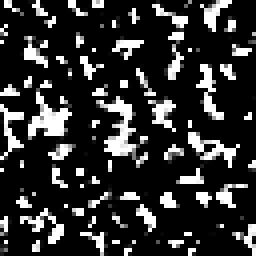} 
\\
\bottomrule
    \end{tabular}
    \caption{Further examples for randomly generated synthetic structures using the trained proposed hybrid models demonstrate the diversity of the trained model.}
    \label{tab:further:generated:samples}
\end{table}

\section{Conclusions and Outlook}
\label{sec:summary}
Generative adversarial networks (GANs) are a promising class of machine learning~(ML) algorithms for microstructure reconstruction.
Current challenges for applying GANs to MCR regarding 2D-to-3D reconstruction and high-resolution are \textit{(i)} the interpretability of the latent space and \textit{(ii)} the applicability to small data sets.

This work presents DA-VEGAN, a hybrid architecture of a $\beta$-variational autoencoder (VAE) enhanced GAN with differentiable data augmentation, to address these issues.
In utilizing the decoder of a $\beta$-VAE as generator of the GAN, DA-VEGAN harnesses the benefits of both ML models.
The $\beta$-VAE introduces a parameter $\beta$ for a flexible and effective penalization of latent space nonlinearities, enabling a potentially smoother and  more interpretable representation of the encoded image.
This facilitates establishing structure-property linkages and simultaneously exploring the space of possible microstructures by means of a well-behaved manifold.
While this already constitutes a major benefit for practical application, still, a major difficulty in GAN- or ML-based reconstruction is the need for a rather large set of data.

On the contrary, in most situations, only a reconstruction algorithm that requires little to no training data is practically useful.
Data augmentation is a common approach for solving this problem by, e.g., translating or rotating images.
However, a naive implementation causes the generative model to imitate the augmented data set with now potentially unrealistically modified images, deteriorating its performance. 
Especially in the context of processed micrograph data, any data modification might alter the characteristics and quality of the image data.
Inspired by the success of differentiable data augmentation in overcoming this issue for pure GAN architectures, an extension to the present hybrid model is developed.
To that end, profound adaptions of the loss functions for the VAE part are implemented, as no existing differentiable augmentation is known to the authors. 

The introduced model architecture and data augmentation algorithm are validated by means of various numerical experiments, whereby excellent results are achieved.
The application of DA-VEGAN for reconstructing simple microstructures of a single elliptical inclusion demonstrates its capability of generating a well-conditioned latent space \emph{and} good reconstruction results by choosing an appropriate trade-off parameter $\beta$.
The impressive performance on extremely small data sets of just one image is shown for 10 different microstructures. In this context, it is emphasized that many recent methods address small data sets by pre-training and fine-tuning, whereas the presented method trains a model \emph{from scratch} despite little data.

Focusing on the concept of a hybrid network, DA-VEGAN is currently limited to 2D, intended for two-phase microstructures and applied to images of resolutions equal or lower than $64\times64$. 
For this reason, future research might focus on \textit{(i)} extending the architecture to 2D-to-3D reconstruction, \textit{(ii)} increasing the microstructure image resolution and \textit{(iii)} enabling the model to reconstruct not only phases, but also crystallographic texture. This would allow for an even broader applicability of DA-VEGAN.

To conclude, a capable and novel ML model for microstructure reconstruction, DA-VEGAN, is presented. Practically relevant, it features a controllable smoothness of the latent space, suitable for structure-property linkages. Most importantly, DA-VEGAN can reconstruct images based on extremely small data sets. It thereby overcomes one of the largest obstacles for the application of ML in materials design -- the need for large sets of data.

\section*{Acknowledgements}
The group of M. Kästner thanks the German Research Foundation DFG which supported this work under Grant number KA 3309/18-1.
The authors are grateful to the Centre for Information Services and High Performance Computing [Zentrum für Informationsdienste und Hochleistungsrechnen (ZIH)] TU Dresden for providing its facilities for high throughput calculations.

\section*{Data availability}
The data to reproduce the presented results is available from the authors upon reasonable request.

\section*{Competing interests}
The authors declare no competing interests.

\section*{Author contributions}
\textbf{Yichi Zhang}: Formal analysis, Investigation, Methodology, Software, Validation, Visualization, Writing - original draft, Writing - review and editing.
\textbf{Paul Seibert}: Conceptualization, Methodology, Supervision, Visualization, Writing - original draft, Writing - review and editing.
\textbf{Alexandra Otto}: Formal analysis, Investigation, Methodology, Software, Writing - review and editing.
\textbf{Alexander Raßloff}: Conceptualization, Methodology, Supervision, Writing - review and editing.
\textbf{Marreddy Ambati}: Conceptualization, Methodology, Supervision, Writing - review and editing.
\textbf{Markus Kästner}: Conceptualization, Funding Acquisition, Supervision, Writing - review and editing.



\printbibliography

\end{document}